\documentclass[journal]{IEEEtran}
\ifCLASSINFOpdf
\else
\fi

\usepackage{amssymb,url,amsmath}
\usepackage{algorithm}
\usepackage{graphicx}
\usepackage{epstopdf}
\usepackage{color}
\usepackage{ifpdf}
\usepackage{url}
\usepackage{enumerate}
\usepackage{array}
\usepackage{flushend}
\usepackage{multirow,booktabs}
\usepackage[pdfstartview=]{hyperref}
\usepackage{mdwlist}
\usepackage[utf8]{inputenc}
\usepackage{todonotes}
\usepackage[classicReIm]{kpfonts}
\usepackage{graphicx} 
\usepackage{fancyhdr}
\usepackage{float}
\usepackage{subcaption}
\usepackage{microtype}
\usepackage{verbatim}
\usepackage{tikz}
\usepackage{ragged2e}
\usetikzlibrary{shapes,arrows,shadows}

\usepackage{tcolorbox}
\newtcolorbox[auto counter]{algorithmbox}[2][]{colback=red!5!white,colframe=red!75!black,fonttitle=\bfseries, title=\alg\thetcbcounter: #2,#1}

\newcommand{\alg}{Algorithm~}

\graphicspath{{images/}}

\IEEEoverridecommandlockouts

\begin{document}
%
\title{Learning Transferable Push Manipulation Skills in Novel Contexts}
%
%
%


\author{Rhys Howard${}^{1}$, Claudio Zito${}^{1,2,\dagger}$
\thanks{$^{\dagger}$Corresponding author {\tt\small  Claudio.Zito@tii.ae}}
\thanks{$^{1}$Intelligent Robotics Lab, School of Computer Science, University of Birmingham, B15 2TT, Birmingham, United Kingdom}
\thanks{$^{2}$Technology Innovation Institute, Abu Dhabi, UAE}
}%

%
%

\markboth{Submitted to IEEE Transactions on Robotics, 2020}%
{Howard \MakeLowercase{\textit{et al.}}: Learning Transferable Push Manipulation Skills in Novel Contexts}
%


\IEEEspecialpapernotice{\copyright 220xx IEEE.  Personal use of this material is permitted.  Permission from IEEE must be obtained for all other uses, in any current or future media, including reprinting/republishing this material for advertising or promotional purposes, creating new collective works, for resale or redistribution to servers or lists, or reuse of any copyrighted component of this work in other works.}

\maketitle

\begin{abstract}
This paper is concerned with learning transferable forward models for push manipulation that can be applying to novel contexts and how to improve the quality of prediction when critical information is available. We propose to learn a parametric internal model for push interactions that, similar for humans, enables a robot to predict the outcome of a physical interaction even in novel contexts. Given a desired push action, humans are capable to identify where to place their finger on a new object so to produce a predictable motion of the object. We achieve the same behaviour by factorising the learning into two parts. First, we learn a set of local contact models to represent the geometrical relations between the robot pusher, the object, and the environment. Then we learn a set of parametric local motion models to predict how these contacts change throughout a push. 
The set of contact and motion models represent our internal model. By adjusting the shapes of the distributions over the physical parameters, we modify the internal model's response. Uniform distributions yield to coarse estimates when no information is available about the novel context. We call this an unbiased predictor. A more accurate predictor can be learned for a specific environment/object pair (e.g. low friction/high mass), called a biased predictor.    
The effectiveness of our approach is demonstrated in a simulated environment in which a Pioneer 3-DX robot equipped with a bumper needs to predict a push outcome for an object in a novel context, and we support those results with a proof of concept on a real robot. We train on two objects (a cube and a cylinder) for a total of 24,000 pushes in various conditions, and test on six objects encompassing a variety of shapes, sizes, and physical parameters for a total of 14,400 predicted push outcomes. Our experimental results show that both biased and unbiased predictors can reliably produce predictions in line with the outcomes of a carefully tuned physics simulator.

\end{abstract}


%
\IEEEpeerreviewmaketitle

\section{Introduction}
Modelling push manipulation so that the outcome of a push can be accurately predicted remains largely an open question, especially in novel situations, e.g. previously unseen objects or same objects in different environments (i.e. a cube on a carpet or on an icy surface). However, as robot make their way out of factories into human environments, outer space, and beyond, they require the skill to manipulate their environment in unforeseeable circumstances. These skills become even more critical to robots encountering conditions as extreme as abandoned mines \cite{thrun2004mines}, the moon \cite{king2016nonprehensile}, or for rescue missions as for the Fukushima Daiichi Nuclear Power Plant.
\begin{figure}[t]
	\centering
	\includegraphics[width=0.95\columnwidth]{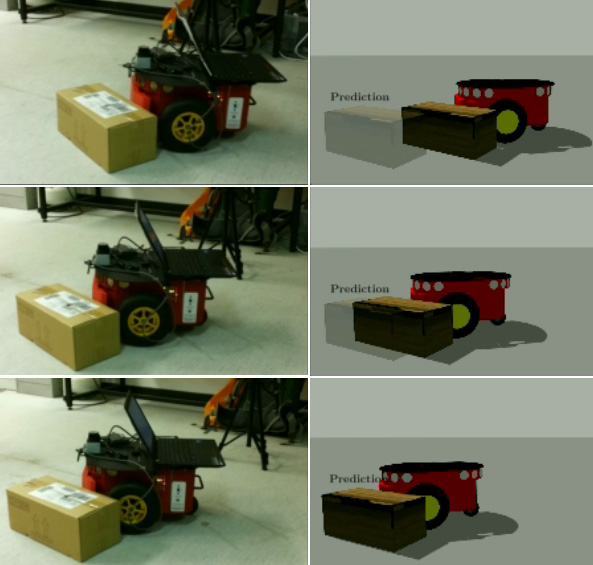}
	\caption{\footnotesize Our  test  scenario.  A  Pioneer  3DX  equipped  with  a  3D  printed bumper pushing a novel object. The robot is capable of predicting the effects of the push even if the carton box is not present in the training data. Left: the real robot executing the action. Right: the action is repeated in simulation where our prediction for the pose of the object at the end of the pushing action is shown.}
	\label{fig:test_scenario}
\end{figure}

Humans possess an internal model of physical interactions \cite{berthoz2002brains,flanagan2006control,flanagan2003prediction,johansson1992sensory,mehta2002forward,witney2000learning} that allows for them to predict the outcome of a physical interaction such as grasping, lifting, pulling, or pushing. Such internal models are the result of an accumulation of a life of physical interactions as opposed to an inherent understanding of physics. Internal models play a crucial role in our ability to manipulate objects, although accuracy is not their main characteristic but to provide a reasonable, adaptable guess. Prediction errors are then dealt with by adjusting on-the-fly our actions. 
Previous efforts in robot pushing have investigated how to design or learn a functional mapping between objects, environments, and motion actions that could be used to make predictions or derive controllers. Mandatory evaluation strategies for these models include accuracy in the predictions or success rates in a chosen scenario but, yet, there is no evidence that the reported performance is in anyway preserved outside the laboratory. While robots in a warehouse can freely navigate and complete tasks such as delivering or checking goods, no system is capable of exploiting push operations for novel objects in novel situations. We hence argue that before we can see an autonomous robot capable of, for example, inserting a box of varied produce onto an over-the-head store shelf, we need to enhance the generalisation capabilities of our models.

Deep learning approaches are appealing on such respect. The recent PushNet have shown the potential of recurrent networks in terms of generalisation to novel objects for re-position and re-orientation tasks using only planar motions. On the opposite note, it is not clear how the recurrent network, trained only on convex objects, is capable of generalising to concave objects, as well as whether it could cope with more complex tasks (e.g. peg-in-the-hole problem) or 3D motions. Finally, it must be noted that PushNet requires $4.8\times 10^5$ training samples over 60 objects to learn planar pushes.   

In contrast, we proposed an intelligible model-based approach for learning a parametric internal model of physical interactions from experience to boost the capability of a robot in predicting the effects of pushing operations in novel contexts. By novel context we refer to objects with different size, shape, or physical properties from the one used in training, as well as different physical properties of the environment. 

In this work we aim to mimic a specific ability observable in humans: the ability of estimating from geometrical properties of the new object/environment where to contact the object for applying the desired push so to obtain a desired (predictable) planar motion of the object. We formulate the problem as follows. First, we learn an internal model (as a forward model) to mimic the casual flow of a push action in terms of the next state of the pushed object given the current state of the system (i.e. object/environment) and the motor commands (i.e. desired push action). Second, we condition the learning process on the observable local contacts to boost the generalisation capabilities of the system; intuitively making predictions on familiar grounds yields to better performance. 

The validity of using local contacts for boosting generalisation has been already demonstrated for robot grasping applications~\cite{kopicki2016one}. However, the grasp synthesis can be considered as a static problem, in which we only aim to compute displacements of the robot's fingers w.r.t. to a target object in a quasi-static fashion to disregard any dynamics. Our previous preliminary investigation in~\cite{stuber2018feature} has shown for the first time that such a formulation can be applied to more dynamical tasks as pushing objects.  
In more details, our internal model is learned by constructing a set of local experts, and each expert encapsulates a part of our understanding of the physical interaction. Each expert is constructed as a probability density function (pdf). We learn two types of experts: i) \textit{contact models}, which learn the local geometrical relations between two bodies, e.g. the contact between the robot pusher and the object to be pushed, or between the object and the environment, and ii) \textit{motion models}, which learn the motion of \textit{contact frames} throughout a push. Each motion model is conditioned on the initial (local) contact frames provided by the contact models, thus enabling the system to make predictions on a familiar ground even when the new object has a different (global) shape from the one used to learn the models.


To learn a contact model we do not require a \textit{Computer Aided Design} (CAD) model of the object. Our objects are acquired via a depth camera, and we refer to them as \textit{Point Cloud Object Models} (PCOMs).
Although our predictors are constructed by using only geometrical features extracted from a point cloud, the motion models are parametric w.r.t. critical physical parameters. The parametric space is represented as two independent pdfs: one for the mass distribution of the object to be pushed, and one for the friction parameter. During training we learn the motion models by repeating the same pushing action with different physical properties drawn from the parametric space distributions. By varying physical parameters, we observe how an object behaves under different conditions. By changing the shape of the pdfs over the parametric space at training time, we bias the predictors to be specialised in a specific context, i.e. low mass/high friction. 

The experimental evaluation aims to demonstrate the generalisation capabilities of the proposed method in novel contexts. Thus we present an extensive set of experiments in a virtual environment as well as a prove of concept on a real Pioneer 3DX robot. The performance of our models are evaluated against the prediction of a carefully calibrated physic simulator.

The experimental results show that the internal model can select a reliable initial contact for the robot to apply the desired push. It can also learn to estimate the initial pose of the object to push, which is critical for estimating the resting pose of the object after the push as a rigid body transformation. It is capable of estimating planar motions of novel shaped objects without knowing the physical parameters, and of improving its performance when some information about the contexts is available in terms of mass and friction distributions.

The remainder of the paper is structured as follows. First a section on related work providing an overview of the history of push manipulation modelling. For a more detailed survey on robot pushing see \cite{stuber_2019}. The next section will describe the background of the model, laying the ground work for the following section which covers the architecture of transferable push manipulation models. An additional section covers our approach to biasing the models during training upon physical parameter distributions. Finally the experimental methodology applied and the ensuing results are then discussed before summarising the conclusions of this paper and proposing directions for future research to consider.
\section{Related Work}
Mason \cite{mason1986mechanics,mason1982manipulator} was the first to start work on constructing a model for the forward modelling of push manipulation motion. This model and those which later built upon it \cite{cappelleri2006designing,flickinger2015performance,lynch1992mechanics,lynch1996stable,mason2001mechanics,peshkin1988motion} are known as analytical models and attempt to closely replicate Newtonian mechanics with their methodology. The main drawback of such approaches is their dependence upon accurate physical parameters and difficulty in modelling friction in some circumstances, as demonstrated by the work of Kopicki and Zito \cite{kopicki2017learning,zito_w2013,zitoIROS2013,zito_2019,zito2012two}.

Recent efforts have instead attempted to build models either partially or entirely built around a set of training data. Zhou et al. \cite{zhou2017fast} proposed a model that combines the underlying structure of an analytical model with a data-driven friction model. Similarly Bauza \& Rodriguez \cite{bauza2017probabilistic} retain some analytically informed structure, but use a data-driven approach based upon \textit{Variational Heteroscedastic Gaussian Processes} to model all the physical processes involved in pushing. In both of these approaches the data-driven nature of the models allows the inherent variance of frictional processes to be captured. This helps to alleviate some inaccuracy seen in predictions from pure analytical models resulting from over idealised friction mechanics.
Further still, Meri{\c{c}}li et al. \cite{mericcli2015push} built a model entirely around a data-driven design. With their approach they made as few assumptions about the physical mechanics involved as possible, relying purely upon collected training data to derive expected motion. This allowed for the model to accommodate for a collection of objects possessing non-quasi-static properties, as these properties were inevitably captured in training data. 

Finally, two neural network based models proposed by Finn et al. \cite{finn2016unsupervised} and Agrawal et al. \cite{agrawal2016learning} respectively attempt to tackle this problem. Finn et al. aimed to utilise video footage alone to generate video footage, however the results of this were both insufficient and ill suited for usage in planning. Meanwhile, the approach Agrawal et al. proposed attempts to simultaneously train both a forward and inverse model for push manipulation motion. Focusing upon the forward model, it first translates images provided to the model into a feature space. From here a relationship between the initial setup in feature space and the motion of the object is established with training data. This in turn allows predictions to be drawn for the motion of the manipulated object across a series of discrete time steps.

Despite the opportunity for autonomous data collection these neural network approaches offer, they also require enormous amounts of data. Given their performance compared to analytical models it may be hard to justify whether the collecting the necessary data is preferential to just obtaining a good set of physical parameter estimates for use with an analytical model alternative. That being said, it appears as though neural network based approaches cannot currently compete with their contemporaries in analytical models and other data-driven models.

Having now established the approaches taken by other contemporary push manipulation models we now proceed to detail our transferable model architecture, beginning with the mathematical and technical background underlying components of the model.

\section{Technical Contributions}\label{sec:technical_contributions}

In  contrast  to  our  previous  work in~\cite{stuber2018feature}, our key technical contributions are as follows:

\textbf{Contact \& Motion Model Selection}:
Previously we trained our models on a single object, i.e. a cube. All the contact models were therefore trained to cope with an object with flat contacts. Predicting motions for a different type of object, i.e. a cylinder, was possible but led to a drop in prediction accuracy. 
In this work we present a method to store several different models trained on different objects and determine which one is most appropriate to use at prediction time. Results of experiments (see Section~\ref{sec:results_selection}) indicate that the method is indeed capable of selecting and applying the most appropriate contact model and motion model based upon principal curvatures associated with the contact model. Prediction accuracy using this method is comparable to the case where contact and motion model are selected by hand using knowledge of the cases to be tested.

\textbf{Pose Estimation}:
In this work we have improved the initial estimate of the object's pose. In previous work a centroid approach was used to approximate the object's initial pose. Here, we show for the first time how to adapt the methodology used by the contact model and re-purpose the \textit{query density} in order to produce an \textit{object position model}. Section~\ref{sec:pose_estimation_accuracy} shows that our proposed model is significantly more accurate than the previously used centroid approach. Furthermore the position model is not limited to just this model and could be used in any situation in which an object's position needs to be estimated from a captured point cloud.

\textbf{Biased Predictors}:
We investigate the effects of biasing vs generalising (i.e. unbiased) models upon coefficient of friction and mass distributions during model training. In our previous work, we assumed the mass of the object to be constant, or sampled from a Dirac distribution, and the friction was sampled from an uniform distribution. In this work, we call these models \textit{unbiased predictors} since we attempt to learn our model over a large range of conditions (i.e. from low to high friction). Our experimental results (Section~\ref{sec:results_accuracy}) show that unbiased predictors are capable of providing a decent prediction when compared to ground truths derived in a physics simulator, i.e. Open Dynamics Engine (ODE). Another benefit of unbiased models is that they offer a greater level of reliability when transferring to novel contexts. Nonetheless, the conclusion of this investigation is that \textit{biased predictors} can be used to offer a significant increase in the accuracy of the motion model and that generalising as it stands typically leads to unintentional biasing.
\begin{table*}[t]
\centering
\caption{\footnotesize List of Symbols by Order of Appearance}
\label{tab:symbols}
\begin{scriptsize}
\begin{tabular}{p{0.05\linewidth} p{0.4\linewidth} | p{0.05\linewidth} p{0.4\linewidth}}
\hline
$O$ & Point Cloud Object Model given as input at training/prediction time. &
$x$ & Surface feature in $SE(3) \times \mathbb{R}^2$. Encapsulates $v$ and $r$. \\
\hline
$v$ & A pose in $SE(3)$. Relative to global origin. Encapsulates $p$ and $q$. &
$r$ & Surface descriptor in $\mathbb{R}^2$. \\
\hline
$SE(3)$ & Standard Euclidean space. The space of poses in 3D, $\{v=(p,q)|p\in\mathbb{R}^3,q\in SO(3)\}$. &
$p$ & Translation vector in $\mathbb{R}^3$. \\
\hline
$SO(3)$ & 3D rotation group of all rotations around the origin of 3-dimensional Eucleadian space, $\mathbb{R}^3$, under the operation of composition. &
$q$ & Quaternion. It describes a rotation in $SO(3)$. \\
\hline
$W^O$ & World origin frame. &
$k$ & A direction of curvature. An orthogonal pair tangential to a surface and perpendicular to its normal can describe the surface's curvature. \\
\hline
$B$ & The object for which the motion resulting from the push operation is modelled. &
$L$ & The robot's manipulative link (e.g. a bumper). \\
\hline
$t_0$ & The time at which the push begins. &
$t_F$ & The time at which the push ends. \\
\hline
$a$ & An action formed of a desired linear velocity $\dot{x}$ and angular velocity $\dot{\theta}$. &
$m$ & The resulting motion from a push in $SE(3)$. \\
\hline
$P(\cdot)$ & Probability density function. &
$c$ & Contact/Position frame in $SE(3) \times SE(3) \times \mathbb{R}^2$. Encapsulates $v$, $r$ and $u$. \\
\hline
$u$ & Relative position representing a relation between two entities in $SE(3)$. Encapsulates $p$ and $q$. &
$E$ & The surrounding environment. \\
\hline
$B^O$ & Origin frame of a given object $O$. &
$h$ & Position of contact/position frame relative to $B^O$, given by $v^{-1} \circ B^O$. \\
\hline
$K(\cdot)$ & Kernel function used in Kernel Density Estimation to approximate a Probability Density Function. Defined by a mean point $\mu$ and bandwidth $\sigma$. &
$\mathcal{N}_n(\cdot)$ & A $n$-variate Gaussian distribution. \\
\hline
$\Theta(\cdot)$ & Distribution formed of a pair of antipodal von Mises-Fisher distributions. &
$d_{p}(\cdot)$ & Translation Distance Function, $L^2$ distance scaled by bandwidth. \\
\hline
$d_q(\cdot)$ & Quaternion Distance Function approximated as $1 \textendash | \langle q_1, q_2 \rangle |$ scaled by bandwidth. &
$d_{r}(\cdot)$ & Surface Descriptor Distance Function, an element-wise computation of the $L^2$ distance scaled by bandwidth. \\
\hline
$D_{{\sigma_{r}}^{-1}}$ & A diagonal matrix formed of the reciprocals of the surface descriptor bandwidths. &
$T$ & Trial rounds $T_p$, $T_q$ and $T_r$ used during KDE to calculate bandwidth multipliers. \\
\hline
$\beta$ & Bandwidth Multipliers $\beta_p$, $\beta_q$ and $\beta_r$ used during KDE to artificially increase bandwidth as necessary. &
$\delta$ & Cut-off distances $\delta_p$, $\delta_q$ and $\delta_r$ used during KDE. \\
\hline
$\alpha_T$ & Rate of Bandwidth Multiplier Change, determines the rate at which $\beta$ decreases with trial rounds. &
$N_K$ & Number of Kernels to be used during Kernel Density Estimation. \\
\hline
$w$ & Scalar kernel weight. &
$X$ & A set of surface features derived from a Point Cloud Object Model $O$. \\
\hline
$\delta_c$ & Contact Model Training Cut-off Distance, determines maximum manipulator contact frame distance during training. &
$\lambda_c$ & Contact Model Training Likelihood Drop-off Rate, determines the rate likelihood decreases as manipulator contact frame distance increases during training. \\
\hline
$w_Z(\cdot)$ & Z Preferencing Function, provides higher likelihood for vertically lower surface features. &
$z(\cdot)$ & Z Accessor Function, provides the Z axis component of a translation $p$. \\
\hline
$w_{CD}(\cdot)$ & Extremity Preferencing Function, provides higher likelihood for surface features at the extremities. &
$centroid(\cdot)$ & Centroid Function, returns the translational centroid of a set of surface features. \\
\hline
$w_{AG}(\cdot)$ & Anti-Grouping Preferencing Function, provides higher likelihood for surface features which would minimise contact frame grouping. &
$C$ & A set of contact frames. \\
\hline
$N_C$ & Number of contact frames. &
$H_{r}$ & Contact Model Selection Surface Descriptor Distance Heuristic, estimates similarity between a Point Cloud Object Model's surface features and a contact model's. \\
\hline
$d_{r}^\prime(\cdot)$ & Near identical to $d_{r}(\cdot)$ except without bandwidth scaling. &
$I$ & The identity matrix. \\
\hline
\end{tabular}
\end{scriptsize}
\end{table*}

\begin{figure*}
	\centering
	\includegraphics[width=0.25\linewidth]{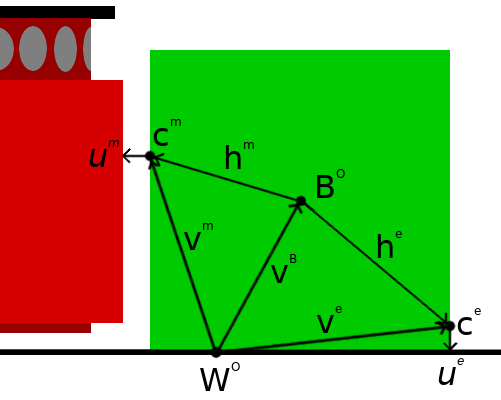}
	\caption{\footnotesize Graphical representation of contact frames. The green area marks the object being manipulated by the robot's manipulator, which is itself the brighter of the reds. The manipulator contact frame is denoted with $c^m$. It is positioned relative to world origin by $v^m$ and its relational pose $u^m$ projects onto the nearest point of the robot's manipulator. Meanwhile an environment contact frame has been denoted with $c^e$. It is positioned relative to the world origin by $v^e$ and its relational pose $u^e$ projects onto the nearest point in the environment. Both the manipulator and environment contact frames have positions relative to the origin of the object $B^O$ given by $h^m$ and $h^e$ respectively. $B^O$ in turn has a position relative to the world origin $W^O$ given by $v^B$.}
	\label{fig:frames}
\end{figure*}

\section{Background} \label{sec:background}
This section provides background information on several of the techniques applied as part of our method. Table \ref{tab:symbols} shows the list of symbols used in this paper at a glance.

\subsection{Surface Features}
Surface features collectively describe the surface of an object and are derived from a 3D point cloud of the object $O$. We define a surface feature as a pair $x=(v,r)\in SE(3) \times \mathbb{R}^2$, where $v=(p,q)\in SE(3)$ is the pose of the surface feature $x$ in the standard Euclidean group defined in a three-dimensional space, and $r\in\mathbb{R}^2$ is a vector in 2 Dimensions (2D) that describes the surface descriptors. We then define $SE(3)$ as $\mathbb{R}^3\times SO(3)$, where $p \in \mathbb{R}^3$ is the translation and $q \in SO(3)$ the orientation of the surface feature. All poses denoted by $v$ are specified relative to a frame at the world origin $W^O$.

To compute the surface normal at point $p$ we use a PCA-based method \cite{kanatani2005geometry}. Surface descriptors correspond to the local principal curvatures around point $p$ \cite{spivak1999geometry}, so that they lie in the tangential plane to the object's surface and perpendicular to the surface normal at $p$. We denote with $k_1 \in \mathbb{R}^3$ the direction of highest curvature, and with $k_2 \in \mathbb{R}^3$ the direction of lowest curvature which is imposed to be perpendicular to $k_1$. Let us define $r = (r_1, r_2) \in \mathbb{R}^2$ as a 2D feature vector to denote the value of the curvatures along directions $k_1$ and $k_2$ respectively. Then, the surface normal and principal directions allow us to define the 3D orientation $q$ that is associated to a point $p$.

\subsection{Rigid Body Motions} \label{subsec:rigid_body_motions}
The push situations we aim to model consist of an object $B$ being pushed by the manipulative link of the robot $L$. The push will begin at time time $t_0$ and finish at time $t_F$, throughout this period of time the robot will attempt to maintain a constant predefined velocity throughout the push, although given the limitations of the robot's dynamics the resulting velocity may be less in cases of high mass and/or friction.

Throughout the push we assume that both $L$ and $B$ are rigid bodies and that the push occurs under quasi-static conditions. Namely, neither $L$ nor $B$ will deform and $B$ will only move as a result of active slow-speed motion as part of the push operation from $L$.

The model is trained for individual predefined actions $a=(\dot{x},\dot{\theta})$ where $\dot{x}$ is the desired linear velocity and $\dot{\theta}$ is the desired angular velocity. These actions together form a complete model which provides a repertoire of actions which can be predicted for during practical usage.

Motion experienced by the object or its parts belongs to $SE(3)$ and is denoted as $m=(p_m,q_m)$, with $p_m$ being the translational component of the motion and $q_m$ the rotational component. For a given action $a$ we aim to train a model that can approximate a Probability Density Function (PDF) over rigid body motions $P(m|a)$.

To achieve this a \textit{Product of Experts} (PoE) approach is used. Rather than model the motion for the object as a whole we sample a set of local surface features to be used as the basis of contact frames. For each of these contact frames an approximation to a PDF $P(m|a,c)$ over rigid body motions for a given action $a$ and contact frame $c$ is derived. The PDF over the object's overall motion can then be predicted by taking the product of the PDFs of each respective contact frame. Each contact frame serves to impose a specific local kinematic constraint on the motion of the object, with each possessing the ability to veto any motion that conflicts with these constraints by returning a probability density of zero for said motion. Therefore an approximation of the global PDF over object motions is derived from a combination of local PDFs over contact frame motions.

The aforementioned contact frames are further decomposed by $c=(v,r,u)$. $v$ and $r$ both correspond to the same components of the underlying surface feature of the contact frame. Meanwhile $u$ belongs to $SE(3)$ and is a relative pose that describes the relation between either $v$ and $L$ or $v$ and the surrounding environment $E$. The first of these will henceforth be described as \textit{manipulator contact frames} $c^m$ while the latter will be described as \textit{environment contact frames} $c^e$.

For each contact frame $c$ there exists a corresponding positioning relative to the origin of the object $B^O$ given by $h$. Meanwhile $B^O$ is itself positioned relative to $W^O$ by $v^B$. Therefore we can derive $h$ from $h = v^{-1} \circ B^O$, where $v^{-1}$ is the inverse pose of $v$ given by $v^{-1} = (- q^{-1} p, q^{-1})$, and $\circ$ is the pose composition operator. The use of $h$ is necessary to relate the PDFs over the local motion of contact frames back to PDFs over the global motion of the overall object.

Further to the description laid out here, all of the aforementioned aspects of the rigid body motion and contact frames are illustrated in Figure \ref{fig:frames}.

\subsection{Kernel Density Estimation} \label{subsec:kde}
In this work \textit{Kernel Density Estimation} (KDE) \cite{silverman1986density} is used to approximate PDFs and is utilised when applying manipulator and environment contact models, position models and motion models.\\
The contact models and position models both utilise a kernel built around surface features. Such a kernel can be described by its mean point $\mu^x = (\mu_{p}^x,\mu_{q}^x,\mu_{r}^x)$ and bandwidth $\sigma^x = (\sigma_{p}^x,\sigma_{q}^x,\sigma_{r}^x)$:
\begin{equation} \label{eq:surface_feature_kernel}
\begin{aligned}
    K^x(x\, &|\, \mu^x,\sigma^x)\ = \\ & \mathcal{N}_3(p^x\, | \, \mu_{p}^x,\sigma_{p}^x)\ \Theta(q^x\, |\, \mu_{q}^x,\sigma_{q}^x)\ \mathcal{N}_2(r^x\, |\, \mu_{r}^x,\sigma_{r}^x)
\end{aligned}
\end{equation}
where $x=(p^x,q^x,r^x)$ is the surface feature being compared against the kernel, $\mathcal{N}_n$ is an $n$-variate Gaussian distribution, and $\Theta$ corresponds to a pair of antipodal von Mises-Fisher distributions forming a distribution similar to that of a Gaussian distribution for $SO(3)$ \cite{fisher1953dispersion}.

In relation to a given surface feature being used as a kernel, $p^x$, $q^x$ and $r^x$ correspond directly to $\mu_{p}$, $\mu_{q}$ and $\mu_{r}$, while $\sigma_x$ is a configurable parameter universal across all kernels in the given model. In the case of both the surface feature acting as a kernel and the one being compared against it, $p^x$ and $q^x$ are sometimes re-positioned from the surface feature itself based off an associated relative pose $u$. This only occurs for manipulator contact models and position models, and does so upon the creation of a query density (See Section \ref{sec:tpmm} for further details). This is because the primary aim of these models is in determining or estimating the positioning of other entities such as the robot or the object's ground truth position.

For our implementation, the previously defined surface feature kernel function is approximated via the use of several distance functions applied over a series of trial rounds, used to track the bandwidth scaling of translational, rotational and surface descriptor components and denoted by $T=(T_p,T_q,T_r)$. Meanwhile the aforementioned approximations and distance functions applied during KDE are as follows:
\begin{equation}
    \mathcal{N}_3(p|\mu_p,\!\sigma_p) \simeq
    \begin{cases}
    0,\!&\!\beta_p d_p(p,\mu_p,\!\sigma_p) \geq \delta_p\\
    e^{-\beta_p d_p(p,\mu_p,\sigma_p)},\!&\!\beta_p d_p(p,\mu_p,\!\sigma_p) < \delta_p
    \end{cases}
\end{equation}
\begin{equation}
    d_p(p,\mu_{p},\sigma_{p}) = \frac{{|| p - \mu_{p} ||}^2}{\sigma_{p}}
\end{equation}
\begin{equation}
    \Theta(q\,|\,\mu_q,\sigma_q) \simeq
    \begin{cases}
    0, & \beta_q d_q(q,\mu_q,\sigma_q) \geq \delta_q\\
    e^{-\beta_q d_q(q,\mu_q,\sigma_q)}, & \beta_q d_q(q,\mu_q,\sigma_q) < \delta_q
    \end{cases}
\end{equation}
\begin{equation}
    d_q(q,\mu_{q},\sigma_{q}) = \frac{1 - | \langle q, \mu_{q} \rangle |}{\sigma_{q}}
\end{equation}
\begin{equation}
    \mathcal{N}_2(r\,|\,\mu_r,\sigma_r) \simeq
    \begin{cases}
    0, & \beta_r d_r(r,\mu_r,\sigma_r) \geq \delta_r\\
    e^{-\beta_r d_r(r,\mu_r,\sigma_r)}, & \beta_r d_r(r,\mu_r,\sigma_r) < \delta_r
    \end{cases}
\end{equation}
\begin{equation}
    d_r(r,\mu_{r},\sigma_{r}) = (r - \mu_{r})^\intercal D_{{\sigma_{r}}^{-1}} (r - \mu_{r})
\end{equation}
where $D_{{\sigma_{r}}^{-1}}$ is a diagonal matrix formed of the reciprocals of the surface descriptor bandwidths $\sigma_{r}$. Meanwhile $\delta = (\delta_p, \delta_q, \delta_r)$ and $\beta = (\beta_p, \beta_q, \beta_r)$ are the cut-off distances and bandwidth scaling parameters respectively. Bandwidth scaling via the parameter $\beta$ is used in the case that a trial round $T$ fails entirely. If a trial round fails, it is because at least one of: $d_p(p,\mu_{p},\sigma_{p}) \geq \delta_p$, $d_q(q,\mu_{q},\sigma_{q}) \geq \delta_q$ or $d_r(r,\mu_{r},\sigma_{r}) \geq \delta_r$ held true in all cases. Bandwidth scaling allows the application of KDE to be reattempted with either the translational, rotational or surface descriptor bandwidth re-scaled. This re-scaling is achieved by incrementing either $T_p$, $T_q$ or $T_r$ and then re-calculating the respective $\beta$ value via the following:
\begin{equation}
\beta = {\alpha_T}^{\textendash T}
\end{equation}
where $\alpha_T$ is a configurable parameter for the rate at which the bandwidth decreases with the increase in trial rounds. Whether to increment $T_p$, $T_q$ or $T_r$ depends primarily upon which trial round value is the smallest and secondarily upon whether $\mathcal{N}_3(\cdot)$, $\Theta(\cdot)$ or $\mathcal{N_2}(\cdot)$ contributed the greatest number of zero likelihoods. This entire process of bandwidth scaling is done under the pretense that even if a contact or position model does not fit a PCOM very well, it still makes sense to provide a tentative result rather than failing completely.\\

Given $N_{K_x}$ surface features from training acting as kernels, a probability density can be derived over surface features in 3D space from the following:
\begin{equation} \label{eq:surface_feature_probability}
    P(x) \simeq \sum_{i=1}^{N_{K}^x} w_i^x K^x(x|x_i,\sigma^x)
\end{equation}
where $x_i$ corresponds to the $i^{th}$ surface feature acting as a kernel and $w_i^x$ corresponds to its weighting with the constraint $\sum_{i=1}^{N_K^x} w_i^x\!=\!1$.\\
As for the motion model, a combination of a contact frame $c$ and its sampled motions $m$ describe a PDF over $SE(3) \times SE(3) \times \mathbb{R}^2$. To account for this a new kernel function can be defined as follows:
\begin{equation} \label{eq:motion_kernel}
\begin{aligned}
    K^m(c,&m\,|\,\mu,\sigma) = \\ & K^c(c\,|\,\mu^c,\sigma^c)\ \mathcal{N}_3(p^m\,|\,\mu_{p}^m,\sigma_{p}^m)\ \Theta(q^m\,|\,\mu_{q}^m,\sigma_{q}^m)
\end{aligned}
\end{equation}
where $m=(p^m,q^m)$ is the translation and rotational motion being compared against the kernel, $c=(v^c,r^c,u^c)$ describes the contact frame the motion is being considered for and $K^c(\cdot)$ is the kernel function for contact frames. Meanwhile $\mu=(\mu^c,\mu^m)$ and $\sigma=(\sigma^c,\sigma^m)$ define the mean point and bandwidth for both the contact frame (in $\mu^c$ and $\sigma^c$) and motion (in $\mu^m$ and $\sigma^m$) components. $\mu^m = (\mu_{p}^m,\mu_{q}^m)$ and $\sigma^m = (\sigma_{p}^m,\sigma_{q}^m)$ further encapsulate the mean point and bandwidth specific to the motion aspect of the kernel. Once again, $\mu_{p}^m$ and $\mu_{q}^m$ correspond to the sampled motion defined by $p^m$ and $q^m$ that is being used as part of the kernel.

The aforementioned $K^c(\cdot)$ closely resembles the surface feature kernel function $K^x(\cdot)$ (Equation \ref{eq:surface_feature_kernel}), and is defined as follows:
\begin{equation}
\begin{aligned}
    K^c(c\, &|\, \mu^c,\sigma^c)\ = \\ & \mathcal{N}_3(p^c\, | \, \mu_{p}^c,\sigma_{p}^c)\ \Theta(q^c\, |\, \mu_{q}^c,\sigma_{q}^c)\ \mathcal{N}_2(r^c\, |\, \mu_{r}^c,\sigma_{r}^c)
\end{aligned}
\end{equation}
where $c=(p^c,q^c,r^c)$ is the contact frame being compared against the kernel, while $\mu^c = (\mu_r^c,\mu_u^c)$ and $\sigma^c = (\sigma_r^c,\sigma_u^c)$ are the mean point and bandwidth of the kernel respectively. From this we can additionally approximate the following PDF:
\begin{equation} \label{eq:contact_frame_probability}
    P(c) \simeq \sum_{i=1}^{N_{K}^c} w_i^c K^c(c|c_i,\sigma^c)
\end{equation}
where $c_i$ corresponds to the $i^{th}$ contact frame acting as a kernel and $w_i^c$ corresponds to the kernel's weighting, once again with the constraint $\sum_{i=1}^{N_K^m} w_i^c\!=\!1$.\\
The main difference between the surface feature kernel and the contact frame kernel is that the surface feature kernel accounts for $v$ while the contact frame kernel does not, instead using $u$. This is because the global placement of the contact frame is irrelevant when determining local kinematic behaviour, while the relative position of the entity relating the contact frame is conversely essential to determining this behaviour.\\
We now have a means of approximating the PDF over contact frames (Equation \ref{eq:contact_frame_probability}) and the motion kernel function for contact frames (Equation \ref{eq:motion_kernel}). It is now possible to apply KDE once more along with conditional probability to derive an estimate of the PDF over motion for a given surface feature with the use of $N_{K_m}$ sampled motions:
\begin{equation} \label{eq:pdf_motion}
    P(m|c,a) \simeq \frac{\sum_{i=1}^{N_{K}^m} w_i^c K_m(c,m\,|\,{c_i, m_i},\sigma)}{P(c)}
\end{equation}
where $a$ is the push action being applied, while $c_i$ and $m_i$ correspond to the contact frame and sampled motion acting as a kernel. In the case of the action being applied $a$, it need not feature in the function itself, as each action effectively has its own model, which combine to provide PDFs over motion for a variety of actions.

\pgfdeclarelayer{background}
	\pgfdeclarelayer{foreground}
	\pgfsetlayers{background,main,foreground}
	
	\tikzstyle{model-part} = [draw, fill=blue!20, text width=6em, 
	text centered, minimum height=4em, drop shadow, fill=red!20, 
	rounded corners, drop shadow]
	\tikzstyle{gazebo} = [draw, fill=blue!20, text width=6em, 
	text centered, minimum height=4em, drop shadow, fill=orange!50, 
	rounded corners, drop shadow]
	\tikzstyle{component} = [draw, fill=blue!20, text width=6em, 
	text centered, minimum height=4em, drop shadow, fill=red!50, 
	rounded corners, drop shadow]
	\tikzstyle{data} = [draw, fill=blue!20, text width=5em, 
	text centered, minimum height=4em, drop shadow, fill=blue!20, 
	rounded corners, drop shadow]
	
\begin{figure*}[t]
\begin{centering}
	\begin{tikzpicture}[thick, scale=0.5, every node/.style={transform shape}]
	\node (south-point) {};
	\path (south-point)+(0,11.5) node (north-point) {};
	\path (north-point)+(-6.5,-1) node (training-text) {\LARGE Training};
	\path (north-point)+(6.5,-1) node (training-text) {\LARGE Prediction};
	\begin{pgfonlayer}{background}
	\path [draw, ultra thick] (south-point) -- node [above] {} (north-point);
	\end{pgfonlayer}
	
	\path (south-point)+(0,3) node (motion-model) [model-part] {Motion Model};
	\path (motion-model.north)+(0,2) node (env-contact-model) [model-part] {Environment Contact Model};
	\path (env-contact-model.north)+(0,1) node (manip-contact-model) [model-part] {Manipulator Contact Model};
	\path (manip-contact-model.north)+(0,1.35) node (position-model) [model-part] {Position Model};
	
	\path (motion-model)+(0,-1.5) node (model-label) {\textbf{Model}};
	\path (env-contact-model)+(0,-1.25) node (contact-model-label) {\textbf{Contact Model}};
	\begin{pgfonlayer}{background}
	\path (position-model)+(2,1) node (model-corner1) {};
	\path (motion-model)+(-2,-2) node (model-corner2) {};
	\path[fill=yellow!20,rounded corners, draw=black!50, dashed]
	(model-corner1) rectangle (model-corner2);
	
	\path (manip-contact-model)+(1.5,1) node (contact-model-corner1) {};
	\path (env-contact-model)+(-1.5,-1.75) node (contact-model-corner2) {};
	\path[fill=green!20,rounded corners, draw=black!50, dashed]
	(contact-model-corner1) rectangle (contact-model-corner2);
	\end{pgfonlayer}
	
	\path (manip-contact-model)+(-5,0) node (stage-one-training-data) [data] {Stage One Training Data};
	
	\path [draw, ->, ultra thick] (stage-one-training-data.east) -- node [above] {} (position-model.west);
	\path [draw, ->, ultra thick] (stage-one-training-data.east) -- node [above] {} (manip-contact-model.west);
	
	\path (motion-model)+(-5,0) node (stage-two-training-data) [data] {Stage Two Training Data};
	
	\path (stage-two-training-data)+(-3,2) node (gazebo) [gazebo] {Gazebo Sim};
	\path (stage-one-training-data)+(-3,0) node (object-robot-descriptors) [data] {Object \& Robot Descriptor};
	\path (stage-two-training-data)+(-3,0) node (smart) [component] {Surface Feature Weighting};
	
	\path [draw, ->, ultra thick] (object-robot-descriptors.south) -- node [above] {} (gazebo.north);
	
	\path [draw, ->, ultra thick] (gazebo.20) -- node [above] {} (stage-one-training-data.240);
	\path [draw, ->, ultra thick] (stage-one-training-data.240) -- node [above] {} (gazebo.20);
	
	\path [draw, ->, ultra thick] (stage-one-training-data.south) -- node [above] {} (stage-two-training-data.north);
	\path [draw, ->, ultra thick] (smart.east) -- node [above] {} (stage-two-training-data.west);
	\path [draw, ->, ultra thick] (manip-contact-model.200) -- node [above] {} (stage-two-training-data.60);
	
	\path [draw, ->, ultra thick] (stage-two-training-data.120) -- node [above] {} (gazebo.340);
	\path [draw, ->, ultra thick] (gazebo.340) -- node [above] {} (stage-two-training-data.120);
	
	\path [draw, ->, ultra thick] (stage-two-training-data.east) -- node [above] {} (motion-model.west);
	\path [draw, ->, ultra thick] (stage-two-training-data.east) -- node [above] {} (env-contact-model.west);
		
	\path (manip-contact-model)+(5,0) node (push-situations) [data] {Push Condition Data};
	\path (push-situations)+(3,0) node (input-data) [data] {Input Data};
	
	\path [draw, ->, ultra thick] (input-data.west) -- node [above] {} (push-situations.east);
		
		\path [draw, ->, ultra thick] (manip-contact-model.east) -- node [above] {} (push-situations.west);
		\path [draw, ->, ultra thick] (env-contact-model.east) -- node [above] {} (push-situations.200);
		
		\path (motion-model)+(5,0) node (predicted-push-outcomes) [data] {Predicted Push Outcomes};
		\path [draw, ->, ultra thick] (position-model.east) -- node [above] {} (predicted-push-outcomes.120);
		
		\path [draw, ->, ultra thick] (push-situations.south) -- node [above] {} (predicted-push-outcomes.north);
		\path [draw, ->, ultra thick] (motion-model.east) -- node [above] {} (predicted-push-outcomes.west);
		
		\path (predicted-push-outcomes)+(3,2) node (query-density) [component] {Query Density};
		
		\path [draw, ->, ultra thick] (query-density.west) -- node [above] {} (push-situations.300);
		\path [draw, ->, ultra thick] (query-density.west) -- node [above] {} (predicted-push-outcomes.60);
		
		\path (predicted-push-outcomes)+(3,0) node (simulated-annealing) [component] {Simulated Annealing Optimiser};
		
		\path [draw, ->, ultra thick] (simulated-annealing.west) -- node [above] {} (predicted-push-outcomes.east);
	
	\end{tikzpicture}
	\caption{\footnotesize Conceptual diagram of the training/prediction pipeline of the model. Training consist of two stages. In the first stage, a point cloud with maximal coverage of the object being trained with is captured. From a combination of this point cloud and information regarding the object/robot the manipulator contact model and position model are generated. In the second stage, the manipulator contact model is used to place manipulator contact frames, while environment contact frames are placed based of a PDF over weighted surface features. Push operations are then repeatedly sampled to gather information regarding the local motion of contact frames. Information regarding the placements of environment contact frames during these pushes forms the basis of the environment contact model, while the observed motion of contact frames forms the basis of the motion model.\\
	Prediction consists of two stages as well, push condition generation and prediction. Push condition generation takes a point cloud of the object to be pushed as input and consists of determining the starting positioning for the robot, as well as the positioning of the manipulator contact frame and environment contact frames. Prediction then consists of using the position model to estimate the object's ground truth position before applying the motion model to the contact frames in order to derive a prediction of the object's motion. The motion can then be combined with the estimated initial position of the object to provide a prediction of the object's final position following the push operation.}
	\label{fig:pipeline}
\end{centering}
\end{figure*}
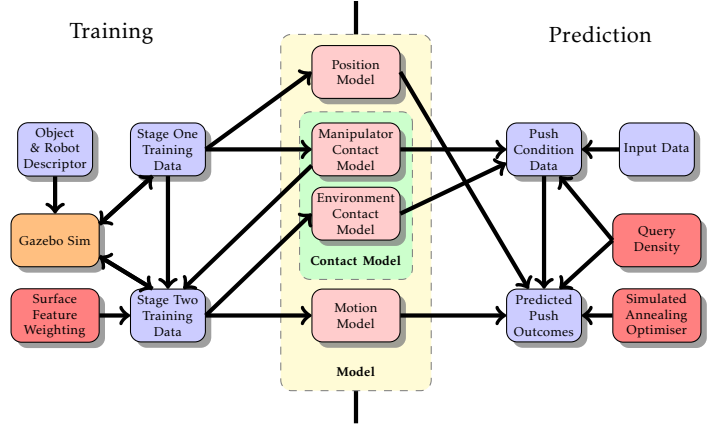
\section{Transferable Push Manipulation Models} \label{sec:tpmm}

This section presents a detailed explanation of our transferable models for push manipulation. Figure \ref{fig:pipeline} provides an overview of the pipeline involved in the training and application of a model.

Our main goal is to to learn how to make predictions on novel contexts, i.e. differently shaped objects that have not been seen in training. We do so by conditioning our predictions in previous experience. We condition on the initial set of contacts, e.g. robot/object (described in Section \ref{subsec:manipulator_contact_model}) and environment/object (Section \ref{subsec:environment_contact_model}) contacts, and on the action to be applied. Under the assumption of rigid bodies, we estimate the initial pose of the object as a reference frame (Section \ref{subsec:object_position_model}) and we infer the object's pose after the action by tracking how the initial contacts have changed. 

When we need to make predictions on a new object, we query the new point cloud to reproduce the contacts seen in training, and to estimate the object's pose (Sections \ref{subsec:query_density} and \ref{subsec:model_selection}). We then apply the motion model for the action we intend to apply (Section \ref{subsec:motion_model}), which yields our final estimate of how the object has moved.

Section \ref{subsec:biasing_upon_friction_and_mass} will present an alternative way for training our motion models by biasing on specific environment/object pairs (e.g. low friction/high mass). 


\begin{figure*}[t]
	\centering
	\begin{tabular}{cc}
	\includegraphics[width=0.25\linewidth]{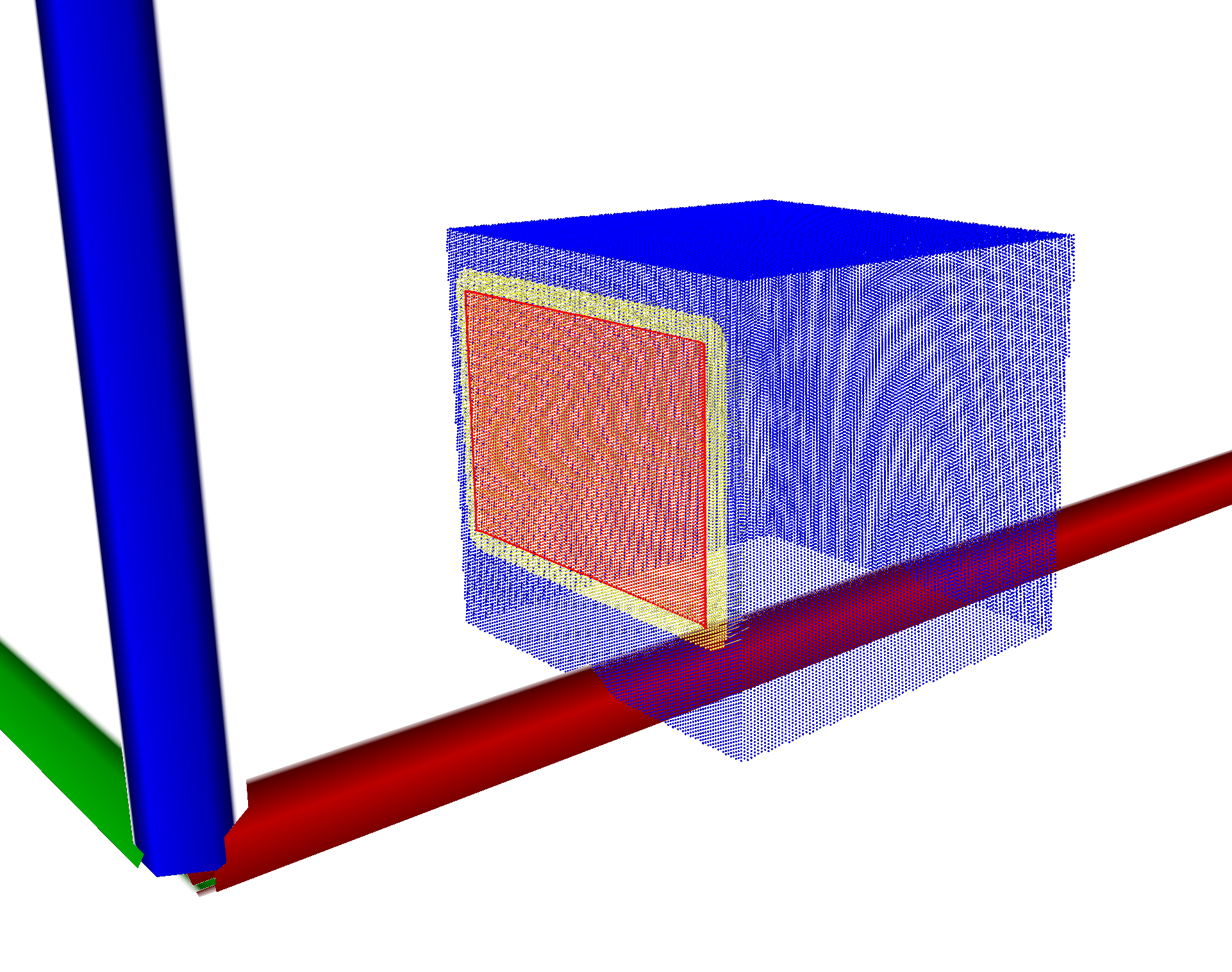} & 
	\includegraphics[width=0.25\linewidth]{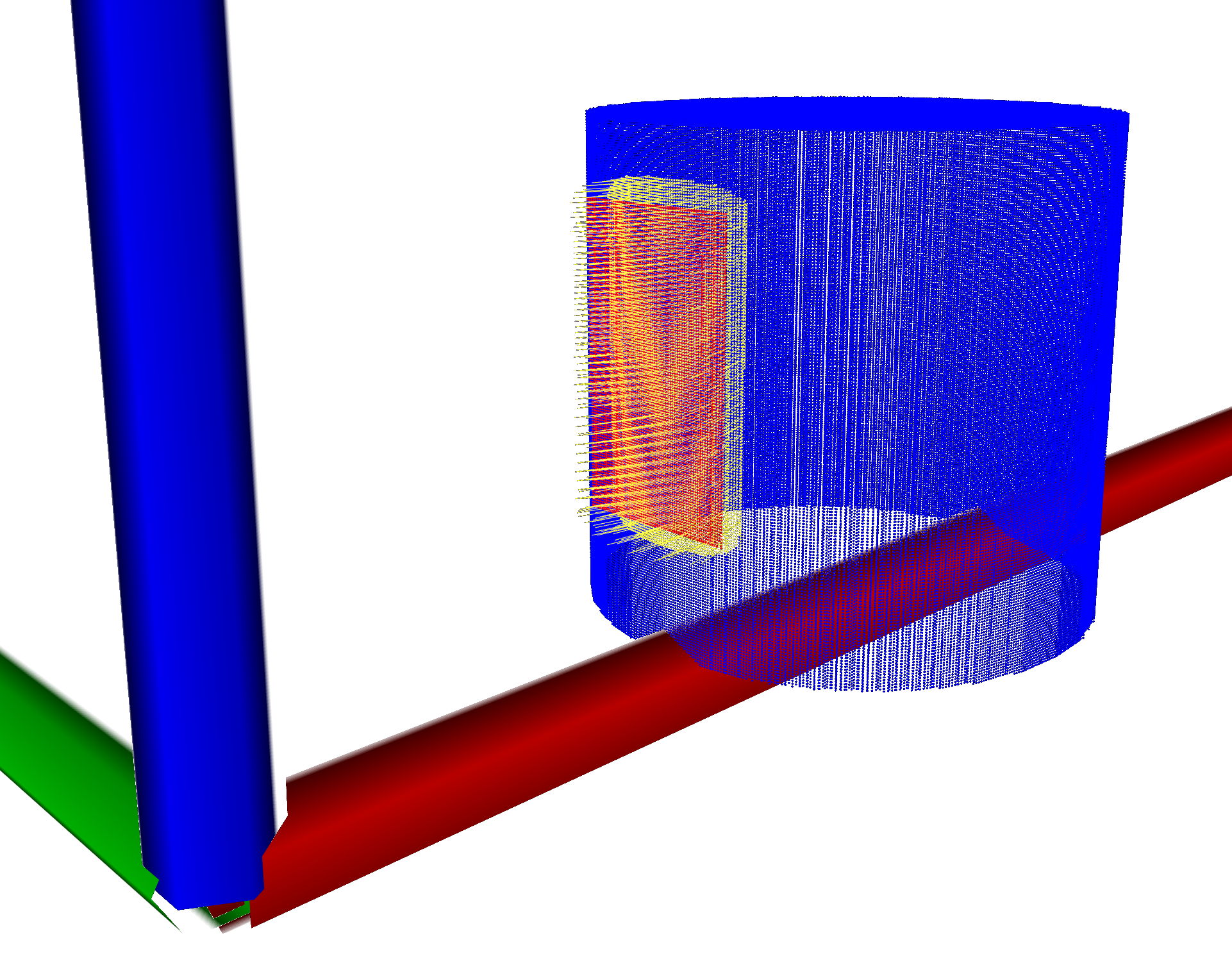} \\
	\end{tabular}
	\caption{\footnotesize Visualisation of the contact models of a cube and cylinder. Blue is the point cloud, yellow are the contacts in the model and red are the projections of the contacts onto the robot's manipulator.}
	\label{fig:push_box_contacts}
\end{figure*}

\subsection{Contact Models}
\subsubsection{Manipulator Contact Model} \label{subsec:manipulator_contact_model}
In order to be able to accurately predict the motion of an object being pushed it is necessary to ensure the placement of the robot should be similar between training time and prediction time. The positioning of the manipulator contact frame corresponds directly with the placement of the robot and so the contact model aims to ensure that the manipulator contact frame is positioned amongst surface features similar to those seen at training time. This is important as initial placement of the robot when pushing has a large impact on the final position of the object and as such prediction time placements should closely reflect training time placements so that accurate predictions are possible.

The contact model consists of a collection of contact frames each denoted $c^m$. These are created by iterating over the surface features $X$ of a training PCOM $O$ and deriving a relational relative pose $u_i^m$ and a weight $w_{i}^c$ for each surface feature $x_i$. The weight for each newly derived contact frame is based on the following calculation:
\begin{equation}
w_{i}^c = \begin{cases}0, & \| u_i^m \| \geq \delta_c\\e^{\textendash \lambda { \| u_{i}^m \| }^{2}}, & \delta_c > \| u_{i}^m \| > 0\\1, & \| u_{i}^m \| \leq 0\end{cases}
\end{equation}
where $\delta_c$ is the cut-off distance for the surface features and $\lambda_c$ is the exponential drop-off rate of the likelihood function. $\| u_i^m \|$ here denotes the smallest distance found by projecting from $x_i$ onto the closest point of each triangular polygon in the mesh of the robot's manipulator. Surface features for which $\| u_i^m \| \geq \delta_c$ would yield a weight of zero and are therefore omitted. Meanwhile $u_{i}^m$ itself is derived from the relative position between the current surface feature and the robot's link $L$. Figure \ref{fig:push_box_contacts} illustrates several contact models superimposed over the PCOMs from which they were derived.

\subsubsection{Environment Contact Model} \label{subsec:environment_contact_model}
Much like the manipulator contact model's purpose is to position the manipulator contact frame, the environment contact model's purpose is to position environment contacts and in doing so best represent the physical constraints of the environment upon the object being manipulated.

Like the manipulator contact model, the environment contact model is comprised of several contact frames each denoted as $c^e$. Just as a manipulator contact $c^m$ has its relation with $L$ described by relative pose $u^m$, an environment contact $c^e$ has a relation with $E$ described by a relative pose $u^e$. Once again, the environment contact frames are constructed around surface features present in a given PCOM. When constructing environment contact frames the assumption is made that $u^e$ always corresponds to the position of the closest point of the floor to the surface feature relative to $v^e$. However, unlike the manipulator contact model, the environment contact model relies upon \textit{Push Data Records} (PDRs) with information regarding the placement of environment contact frames during training of the motion model, which is itself also composed of PDRs.\\
When training the environment contact model the contact frames must be placed based upon intuition of what constitutes a good placement for environment contact frames. Since we assume the closest point from anywhere on the object will always be the floor, it makes sense to preference placing environment contact frames close to the floor. Therefore given the surface features $X$ of a PCOM $O$ we construct a function for weighting a given surface feature $x$ as follows:
\begin{equation}
w_Z(x,X) = e^{-\frac{z(p^x)}{\max_{x_i \in X} z(p_i^x)}}
\end{equation}
where $z(\cdot)$ is a function that returns the Z component of the input translation. This function provides a greater weighting to surface feature translations with lower Z values, therefore making it more likely that surface feature translations closer to the ground plane will be selected.

Another aspect to consider when placing contact frames is that surface features at the outermost extremities of an object are much more likely to come into contact with the environment. For example, if we consider a cube, no part of the cube can come into contact with the environment without at least one vertex of the cube also coming into contact with the environment. With this in mind we define a second weighting function for a surface feature $x$:
\begin{equation}
w_{CD}(x,X) = \frac{{|| p^x - centroid(X) ||}^2}{\max_{x_i \in X}{|| p_i^x - centroid(X) ||}^2}
\end{equation}
where $centroid(\cdot)$ is a function that returns the centroid formed by the translation components of the input set of surface features. This function makes it more likely that environment contacts will be sampled from surface feature translations at the outermost extremities of the object as determined by distance to the PCOM's centroid.

The final aspect that must be considered when placing contact frames is to avoid placing environment contact frames close to one another. If certain regions of surface features are weighted particularly highly then it raises the likelihood of several environment contact frames being placed here during training. Since the motion model trains its own model based upon the placement of contact frames during training, it is important to ensure a diverse placement of environment contact frames in order to best learn kinematic constraints relevant to all surface features present. With this in mind, a final weighting function for a given surface feature $x$ is defined:
\begin{equation}
w_{AG}(x,X,C^e) = \prod_{c_i \in C^e}^{N_C}\frac{{||p - p_i^{c}||}^2}{\max_{x_j \in X} {|| p_j^x - p_i^{c} ||}^2}
\end{equation}
where $C^e$ are the environment contact frames that have been placed thus far. This function lowers the weighting of surface feature translations near the environment contacts sampled so far hence reducing the likelihood of environment contact grouping occurring.

Having defined the various surface feature weighting functions, we can now define a PDF over surface features:
\begin{equation}
P(x \, | \, X, C^e) = w_Z(x,X) \, w_{CD}(x,X) \, w_{AG}(x, X, C^e)
\end{equation}
Using this PDF we can sample surface features from which environment contact frames can be constructed. In order to do this $u^e$ must be derived. We assume the floor is always the closest point in the environment, for a surface feature $x$ with translation $p=(x,y,z)$, we find $u^e$ to be $(0,0,-z)$ since the closest point on the floor is $z$ directly below the contact frame. Following the placement of all environment contact frames for a given training push the setup of environment contact frames will be recorded in training push's corresponding PDR.

Unlike the manipulator contact model or the position model at prediction time a query density is not used in placing the environment contact frames as prediction time. Instead we sample from a uniform distribution of surface features derived from the input PCOM and use the resulting surface feature to construct a candidate environment contact frame. Equation \ref{eq:contact_frame_probability} is then used with $N_K^c$ environment contact frames acting as kernels to provide a likelihood value for this candidate environment contact frame. This process is repeated for a predefined number of iterations before accepting the candidate with the highest likelihood to become an environment contact frame. The entire process is then repeated several times to provide a predefined number of environment contact frames to be utilised by the motion model in its predictions.

\begin{figure*}[t]
	\centering
	\includegraphics[width=0.3\linewidth]{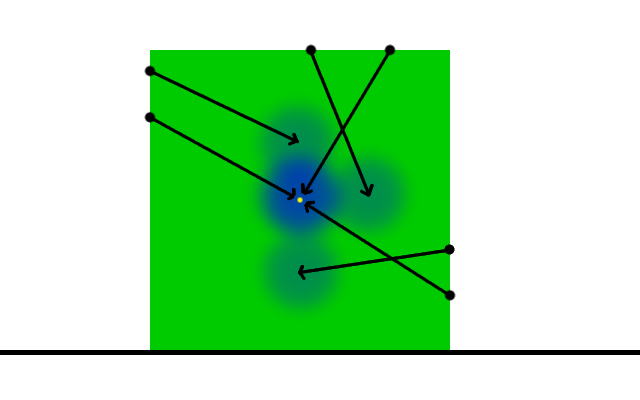}
	\caption{\footnotesize Illustration of the position model application process. The ground truth position of the object demarked by green is indicated with a yellow point. The query density kernels are represented here in blue. These kernels form a PDF within $SE(3)$ space over the object's position. As such, the kernels can be supplied to an simulated annealing based optimiser in order to estimate the position of the object.}
\end{figure*}

\subsection{Object Position Model} \label{subsec:object_position_model}
The purpose of the object position model is to estimate the initial pose of the object to be pushed. To achieve this a structure closely resembling the manipulator contact model is used, except rather than modelling the relationship between surface features and the robot's manipulative link, the relationship between surface features and the object's position is modelled. This requires the definition of a position model made up of position frames in the same way that the contact model is made up of contact frames.

The construction of the position model largely mirrors that of the contact model with a few key differences. Firstly, a position frame is denoted by $c^p = (v^p,r^p,u^p)$ as opposed to $c^m$. More important however is that during training rather than calculating the $u^p$ by calculating the relative position of L, $u^p$ can simply be made to equal $h$ for the position frame, which we have already shown to be calculable as $v^{-1} \circ B^O$. Additionally, in contrast to the contact model the weight of a position $w_i$ for a surface feature $x_i$ is calculated as follows:
\begin{equation}
w_i = d_{r}(r_i, \Bar{r}, \sigma_r^x)
\end{equation}
where $r_i$ is the surface descriptor for the surface feature, $\Bar{r}$ is the mean surface descriptor given by the set of surface features $X$ present in the PCOM $O$ provided at training time. As such the weight of the position frame increases the further the surface descriptors of the position frame are from the mean value of the surface descriptors. This is desired because it means that features that are likely to appear less often on the object will be weighted higher as they act as better indicators of the object's position. For example, if we consider a cube the presence of a vertex provides the exact location of the centre of the cube relative to the position and orientation of the vertex. Meanwhile the centre of the cube could be in a multitude of positions relative to the position and orientation of a flat surface upon the cube, as there are countless flat surfaces across a cube. In other words, salient surface features are implicitly less likely to have a large number of potential relative positions for the object's centre and therefore ought to be afforded a greater weighting.

\begin{figure*}
\centering
\includegraphics[width=0.6\linewidth]{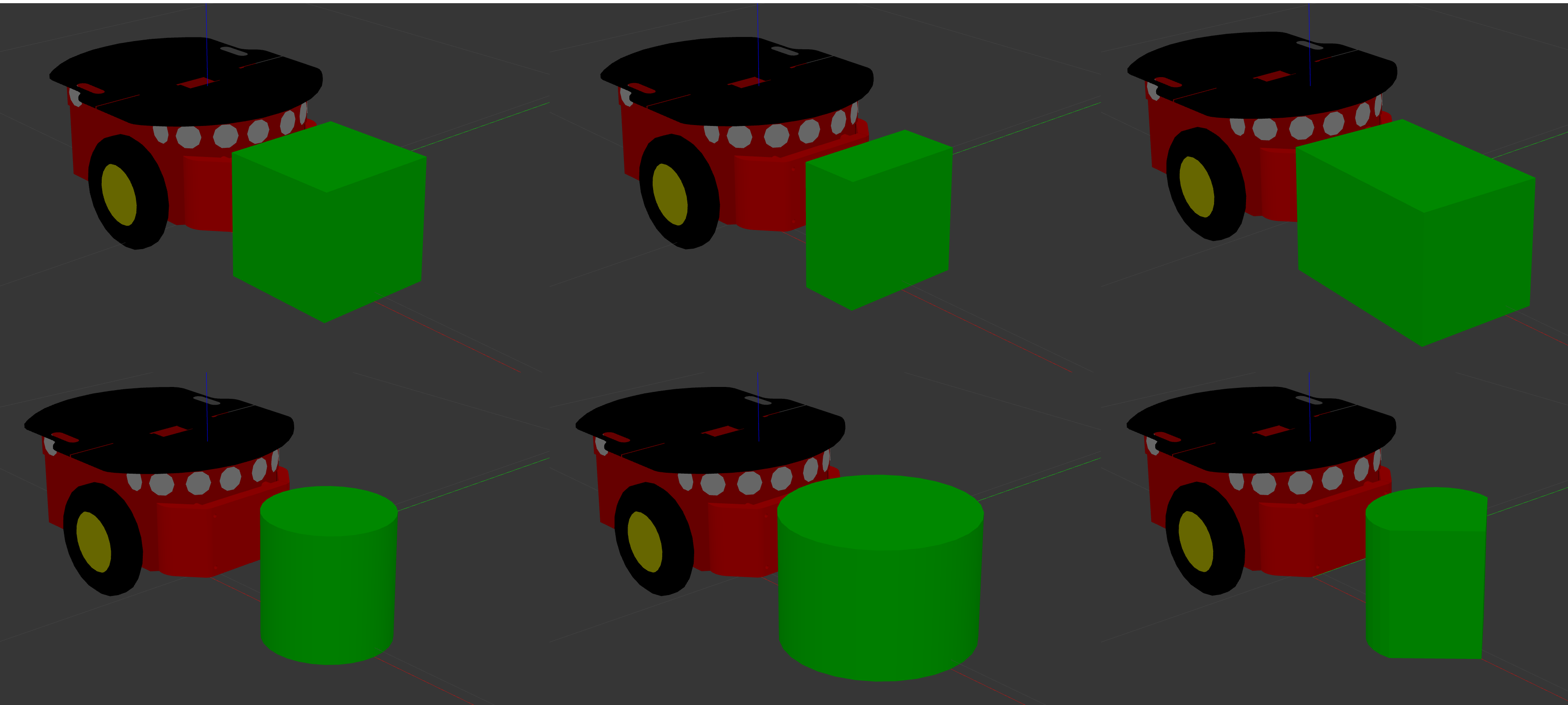}
\caption{\footnotesize An example of the variety of objects that a given model can be applied to. Moving clockwise from top left: a cube with sides of length $20\ cm$, a rectangular prism with dimensions $10\ cm \times 20\ cm \times 20\ cm$, a rectangular prism with dimensions $30\ cm \times 20\ cm \times 20\ cm$, a cylinder of height $20\ cm$ and radius $10\ cm$, a cylinder of height $20\ cm$ and radius $15\ cm$, and a hybrid object formed of half a cylinder and a triangular prism amounting to dimensions of $20\ cm \times 20\ cm \times 20\ cm$.}
\label{fig:objects}
\end{figure*}

\subsection{Query Density} \label{subsec:query_density}
The query density represents a series of distributions resulting from the combination of a learnt model with a PCOM. The object is represented as a partial PCOM of the visible object's surface. Through this process two quantities are estimated. By applying the manipulator contact model described in Section \ref{subsec:manipulator_contact_model}, we estimate the placement of the robot's link w.r.t. local surface features such that the contact is as similar as possible to the training examples. Since several manipulator contact models can be learned, as shown by Figure \ref{fig:push_box_contacts}, we simply query the new object with all of the available models and select the one that produces the best match with the contacts seen in training. A detailed explanation of our selection process is described in Section \ref{subsec:model_selection}. 
Additionally we use the object position model (Section \ref{subsec:object_position_model}) to determine a reference frame for the visible point cloud. Once these three quantities are computed, we are ready to apply the motion model (Section \ref{subsec:motion_model}) to estimate the effect of a pushing action on the novel object.
This process is designed to be transferable such that a model trained upon a single object can be applied to a variety of previously unseen objects such as those illustrated in Figure \ref{fig:objects}.
In order to be able to derive a PDF over robot/object positions that is more representative of the presented PCOM $O$ a new set of kernels must be derived. To do this, a surface feature $x$ is first sampled from a uniform distribution of the surface features $X$ of PCOM $O$.
A weight $w_{i,j}^n$ is then derived for each contact or position frame in the corresponding model in preparation for the next step:
\begin{equation}
w_{i,j}^n = w_j^c d_r(r^x , r_j^c, \sigma_r^c)
\end{equation}
where $w_j^c$ and $r_j^c$ are the weight and surface descriptors of the $j^{th}$ contact frame, and $r^x$ is the surface descriptor of $x$.\\
Having assigned a neighbourhood weight $w_{i,j}^n$ to each contact/position frame, a frame is sampled from a non-uniform distribution over the contact/position model. The likelihood of a frame being selected directly corresponds to the neighbourhood weight it was assigned in the last step. Once a contact/position frame has been selected, a new frame is constructed using the $v$ and $r$ components from the sampled surface feature $x$ and the $u$ component from the sampled contact/position frame. Finally another weight ${w_i^c}^\prime$ is calculated which will act as the new weight to be associated with the newly constructed contact/position frame:
\begin{equation} \label{eq:qd_kernel_weight}
    {w_i^c}^\prime = \sum^{N_K^c}_{j=1} w_{i,j}^n
\end{equation}
where $N_K^c$ corresponds to the number of frames belonging to a manipulator contact or position model. This entire process is then repeated until a desired number of kernels have been created.\\
Once all of the kernels have been created, a KDE approximation of a PDF over robot/object positions can be derived. Because we want to ultimately produce candidates for the position of the robot/object all the kernels are positioned at $u \circ v$. In other words, we shift the kernels from the position of the sampled surface feature based upon the relative pose of the sampled contact/position frame. Furthermore, when approximating the likelihood using KDE during optimisation, the surface descriptor distance is omitted, as this has already been encoded via the weight of kernels calculated by Equation \ref{eq:qd_kernel_weight}. This allows the optimisation to take place purely within $SE(3)$ space.\\
Now having our KDE approximation, we first sample from a discrete distribution over the mean points of the kernels in 3D space, using the weights associated with each kernel to determine their likelihood of being selected. Having established several candidates we then perform simulated annealing based optimisation upon the candidates aiming to maximise the likelihood approximated by KDE. Once this process is complete, we take the candidate with the highest likelihood and use it either to determine the starting position for the robot or an estimate for the position of the object being pushed. Additionally, in the case of the manipulator contact model, the kernel with the closest mean point to the candidate with the highest likelihood will also be returned. Since the kernels are formed of contact frames, this contact frame will then become the manipulator contact frame to be used as part of the motion model.

\subsection{Motion Model} \label{subsec:motion_model}
The motion model consists of a series of PDRs containing information regarding the local motion of contact frames during training pushes. These motions are then combined with the contact frames that have been placed at prediction time to create a KDE approximation to a PDF over final object positions for each of the contact frames.
Prior to any KDE approximation taking place a weighting $w_i^P$ for each PDF is derived. For environment contact frames $w_i^P$ is 1. Meanwhile the manipulator contact frame PDF weight $w_i^P$ is calculated as follows:
\begin{equation}
w_i^P = \frac{N_K^{c^e}}{N_K^{c^m}}
\end{equation}
where $N_K^{c^e}$ is the number of environment contact frame kernels and $N_K^{c^m}$ is the number of manipulator contact frame kernels.\\
Besides the PDF weightings, the weightings of individual contact frame kernels $w_i^c$ are theoretically uniform across all kernels by default. In practice however, the contact frame kernels are used to encode the $K^c(c|\mu^c,\sigma^c)$ and $P(c)$ parts of Equation \ref{eq:pdf_motion}. This again, allows optimisation to be carried out over $SE(3)$ space and can be done since the contact frames have already been placed.\\
Finally, because we wish to optimise final object position candidates using a PoE technique the kernels need to be represented as global object motions rather than as local contact frame motions. To do this each approximated PDF is shifted based upon the position of its contact frame relative to $B^O$ given by $h$. This shifts all the PDFs into the same motion space. From here a new KDE approximation to a PDF over global object motion for an action $a$ can be defined using the PoE technique discussed previously:
\begin{equation}
P(m|a) \simeq \prod_{i=1}^{N_C} w_i^P P(m|a,c_i)
\label{eq:poe}
\end{equation}
where $N_C$ is the number of contact frames used in prediction and $c_i$ is the $i^{th}$ contact frame.\\
Having defined an approximation of a PDF over global object motion simulated annealing optimisation can once again be applied as previously described in Subsection \ref{subsec:query_density}. The final result of this optimisation will be several candidates for object motion with associated likelihoods. From here these motions can be applied as a transformation to the current object pose at $B^O$, providing a prediction for the final position of the object following the push action $a$.

\subsection{Contact \& Motion Model Selection}\label{subsec:model_selection}
While the model described in this paper is indeed transferable and can produce good results for a variety of unseen objects, objects with significantly different surface features may suffer in prediction accuracy. The underlying problem is twofold, both the physical behaviour of the object is likely to be different and the model itself will not adapt well to being exposed to significantly different surface descriptors. Therefore we overcome this issue by implementing a model library, from which an appropriate model can be drawn and applied to an object presented at prediction time.\\
In order to allow for this it is necessary to introduce a heuristic as part of the manipulator contact frame query density to measure the similarity of the surface features of the point cloud and the contact model used in its creation. The calculation of a feature distance heuristic $H_{r}$ is carried out during the creation of the kernels of the Query Density and is calculated as follows:
\begin{equation}
H_{r} = \sum_{i=1}^{N_K^q} \sum_{j=1}^{N_K^c} \sqrt[4]{d_{r}^{\prime}(r_i^q, r_j^c)}
\end{equation}
where $N_K^q$ is the number of query density kernels being created, $N_K^c$ is the number of manipulator contact model contact frames being used in the kernel creation process and $r_i^q$ is the surface descriptors of the current sampled feature being compared against the surface descriptors of the current contact model contact frame $r_j^c$. $d_{r}^{\prime}(\cdot)$ is a modified version of the surface descriptor distance function, defined as follows:
\begin{equation}
d_{r}^{\prime}(r_1, r_2) = (r_1 \textendash r_2)^\intercal I (r_1 \textendash r_2)
\end{equation}
where $r_1$ and $r_2$ are surface descriptors and $I$ is the identity matrix. The heuristic provides a measure on how similar the surface features observed in the prediction time PCOM are to those that make up the manipulator contact model. A lower value indicates that the contact model used in the creation of the query density has features closer to those perceived in the object's point cloud.

During prediction, several query densities can be produced using the different manipulator contact models available, then the query density with the lowest surface feature distance heuristic is selected and used as per usual. When it comes to the application of the motion model, each trained contact model will be associated with a trained motion model. Therefore the manipulator contact model of the previously selected query density can be used to infer the motion model that ought to be applied in this instance. Once the inferred motion model is applied the outcome of the push will have been predicted using the most appropriate models for the object in question.

\section{Biasing Transferable Push Manipulation Models}
\label{subsec:biasing_upon_friction_and_mass}
The primary novel contribution of this paper is the implementation and analysis of \textit{Physical Parameter Generalising} and \textit{Physical Parameter Biasing} during training. Physical parameters such as friction and mass often require explicit definitions when working with an analytical means of prediction. Hence a naive approach to handling physical parameters would be to set them as fixed values for training based upon a good approximation of physical parameters at prediction time. However, such an approach is prone to overfitting as the inherent variance of physical parameters in the real world is unaccounted for.

Therefore the physical parameters ought to be drawn from a distribution. Generalising and biasing each correspond to a type of distribution from which physical parameter values may be drawn. Generalising describes the use of a uniform distribution over a broad range of values, the aim being to provide competent predictions for a wide variety of objects and physical environments. Meanwhile, biasing describes the use of a narrow normal distribution centered upon predetermined mean value reflective of the expected operating conditions of the robot in question.

It is based upon the theoretical underpinning of these distributions that we hypothesise that models trained upon biased physical parameters will provide a greater degree of accuracy than achievable with a generalised. However, the generalised model is expected provide a reasonable level of accuracy across a wide range of operating conditions. This is in contrast to the biased models, which are only expected to perform to a high degree of accuracy in conditions similar to those described by the parameter values upon which the biased models were trained.


\section{Experimental Setup} \label{sec:methodology}

\tikzstyle{method-step} = [draw, fill=gray!20, text width=6em, 
	text centered, minimum height=4em, drop shadow]
	
\begin{table*}[t]
\centering
\scriptsize
\caption{\footnotesize Experimental parameter descriptions for prediction accuracy experiments.}
\begin{tabular}{|p{2.5cm}|p{5cm}|p{2.5cm}|p{5cm}|}
    \hline
    \multicolumn{2}{|c|}{Contact Model Generation} & \multicolumn{2}{|c|}{Condition Generation} \\
    \hline
    Distance & The cut-off distance $\delta_{m}$ for determining which contacts to keep when generating a contact model. & Number of Conditions to Generate & Number of push conditions to generate. \\[0.75cm]
    Lambda & The exponential drop-off rate $\lambda_c$ used when calculating weights for contacts when generating a contact model. & Number of Environment Contacts & Number of environment contacts to place as part of the process of generating each push condition. \\[0.2cm]
    & & Number of Samples When Generating Environment Contacts & Number of samples to take from the environment contact model when generating each environment contact. \\
    \hline
    \multicolumn{2}{|c|}{Motion Model Training} & \multicolumn{2}{|c|}{Ground Truth Generation} \\
    \hline
    Number of Actions & Number of actions that the motion model will be trained for. & Number of Actions & Number of actions that will be simulated for each push condition. \\[0.2cm]
    Angle Range & Defines the range of angular velocities from which each action will be derived. & Angle Range & Defines the range of angular velocities from which each action will be derived. \\[0.2cm]
    Action Duration & Duration of push operation. & Action Duration & Duration of push operation. \\[0.2cm]
    Action Speed & Target speed of robot during push operation. & Action Speed & Target speed of robot during push operation. \\[0.2cm]
    Samples Per Action & Number of sample push simulations to carry out and record for each action. & Samples Per Action & Number of sample push simulations to carry out and record for each combination of action and push condition. \\[0.2cm]
    Object Mass & Object mass value or distribution from which the object mass will be sampled. & Object Mass & Object mass value or distribution from which the object mass will be sampled. \\[0.2cm]
    Object Coefficient of Friction & Object coefficient of friction value or distribution from which the object coefficient of friction will be sampled. Only present in initial experiments, coefficient of friction parametrisation was moved to ground plane to better represent real world conditions for later experiments. & Object Coefficient of Friction & Object coefficient of friction value or distribution from which the object coefficient of friction will be sampled. Only present in initial experiments, coefficient of friction parametrisation was moved to ground plane to better represent real world conditions for later experiments. \\[0.2cm]
    Ground Plane Coefficient of Friction & Ground plane coefficient of friction value or distribution from which the ground plane coefficient of friction will be sampled. Only present in later experiments as discussed above. & Ground Plane Coefficient of Friction & Ground plane coefficient of friction value or distribution from which the ground plane coefficient of friction will be sampled. Only present in later experiments as discussed above. \\[0.2cm]
    Number of Environment Contacts & Number of environment contacts to be recorded in conjunction with each push simulation. & & \\
    \hline
    \multicolumn{4}{|c|}{Prediction Generation} \\
    \hline
    Number of Environment Contacts & \multicolumn{3}{|p{12.5cm}|}{Number of environment contacts to use for each push condition when predicting the final object transform. Environment contacts are stored as part of each push condition following their placement in push condition generation.} \\[0.5cm]
    Environment Contact Kernels & \multicolumn{3}{|p{12.5cm}|}{Number of environment contact kernels to use for each push condition when predicting the final object transform. Kernels come from the PDRs that comprise the motion model in use.} \\[0.5cm]
    Manipulator Contact Kernels & \multicolumn{3}{|p{12.5cm}|}{Number of manipulator contact kernels to use for each push condition when predicting the final object transform. Kernels come from the PDRs that comprise the motion model in use.} \\
    \hline
\end{tabular}
\label{tab:experimental_parameter_descriptions}
\end{table*}

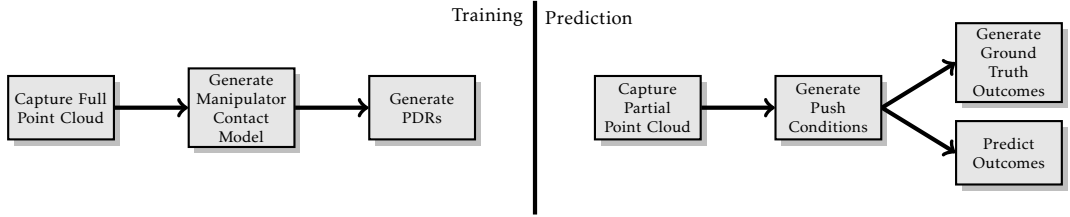
\begin{figure*}[th]
\centering
\begin{tikzpicture}[thick, scale=0.6, every node/.style={transform shape}]
\node (step1) [method-step] {Capture Full Point Cloud};
\path (step1)+(4,0) node (step2) [method-step] {Generate Manipulator Contact Model};
\path (step2)+(4,0) node (step3) [method-step] {Generate PDRs};
	
\path [draw, ->, ultra thick] (step1.east) -- node [above] {} (step2.west);
\path [draw, ->, ultra thick] (step2.east) -- node [above] {} (step3.west);

\path (step3)+(2.5,2.5) node (point1) {};
\path (step3)+(2.5,-2.5) node (point2) {};
\path (point1)+(-1.05,-0.5) node (training-text) {\large Training};
\path (point1)+(1.20,-0.5) node (training-text) {\large Prediction};

\path [draw, ultra thick] (point1) -- node [above] {} (point2);

\path (step3)+(5,0) node (step4) [method-step] {Capture Partial Point Cloud};
\path (step4)+(4,0) node (step5) [method-step] {Generate Push Conditions};
\path (step5)+(4,1) node (step6) [method-step] {Generate Ground Truth Outcomes};
\path (step5)+(4,-1) node (step7) [method-step] {Predict Outcomes};

\path [draw, ->, ultra thick] (step4.east) -- node [above] {} (step5.west);
\path [draw, ->, ultra thick] (step5.east) -- node [above] {} (step6.west);
\path [draw, ->, ultra thick] (step5.east) -- node [above] {} (step7.west);
\end{tikzpicture}
\caption{\footnotesize Illustration of the generalised "Model Prediction Accuracy" experiment methodology.}
\label{fig:prediction_accuracy_methodology}
\end{figure*}

Our experiments were conducted using ROS Kinect and Gazebo Sim 9.0 with the Open Dynamics Engine (ODE). Our test robot is a Pioneer 3-DX mobile robot equipped with a bumper affixed in front of it to provide a flat surfaces for the pushes. The robot was controlled with a ROS package called MoveIt.  

The generalised methodology used during the experiments is illustrated in Figure \ref{fig:prediction_accuracy_methodology}. At training time, a full point cloud of the object to be pushed is acquired from multiple views of a virtual depth camera in Gazebo. The robot is placed with the desired contact between the robot's bumper and the object to learn the manipulator contact model (Section~\ref{subsec:manipulator_contact_model}) and the environment contact model (Section~\ref{subsec:environment_contact_model}). Once the physical critical parameters (i.e. friction coefficient and mass distribution) are sampled from the desired distributions, a push action is generated and a PDR is recorded.

At prediction time, a new point cloud is captured with a single shot of the virtual depth camera, to best mirror typical circumstances in a real application.
The system queries the novel point cloud with the contact model and selects the more appropriate. Then it executes the action associated with that particular contact model. Multiple actions could be associated with a particular contact model, however this goes outside the scope of this paper. We are not interested at this point to move the object in a desired final configuration, our aim is to prove that we can make reasonable predictions of how the object behaves in novel contexts. Our model's predictions are compared against the ODE's outcomes which we assume to be the ground truth. The real values of the physical parameters used by the ODE to perform the action are unknown to our system.
The prediction accuracy measure described in Section \ref{subsec:prediction_accuracy_measure} is used to evaluate our model's performance. Descriptions of experimental parameters used by this methodology are detailed in Table \ref{tab:experimental_parameter_descriptions}.

\subsection{Prediction Accuracy Measure} \label{subsec:prediction_accuracy_measure}

A suitable heuristic of prediction accuracy is necessary to compare the efficacy of various trained models. This measure evaluates the displacement between a predicted pose and the true observed one. We use this to evaluate both the accuracy in the pose estimation model (Section~\ref{subsec:object_position_model}) and the motion model (Section~\ref{subsec:motion_model}). The following heuristic $H_{acc}$ correlates with the error in a prediction, as such a smaller heuristic value is associated with a greater level of prediction accuracy. The heuristic is defined as:

\begin{equation}\label{eq:pam}
H_{acc} = {\|(p_e \textendash p_{gt})^\intercal D_S\|} + \min_{q_{gt} \in Q_{gt}} (1 \textendash {\langle q_e, q_{gt} \rangle}^2)
\end{equation}
where $p_e$ and $q_e$ are the translation and orientation of the predicted final object transform, $p_{gt}$ and $Q_{gt}$ are the translation and orientations of the ground truth object transform and $D_S$ is the reciprocals of the dimensions of the object represented as a diagonal matrix.

Rather a single ground truth orientation being considered, instead $Q_{gt}$ separate orientations are considered and the smallest resulting quaternion distance is utilised. This is to account for the fact that for certain objects there are several orientations that appear identical. For example, a featureless, textureless cube appears identical across 24 orientations for a given translation.

Furthermore the heuristic scales its resulting value based upon the size of the object being manipulated. This decision was taken upon the insight that the size of the object is often reflective of the accuracy required for a task. For example, a centimeter when moving a crate is relatively insignificant compared to a centimeter when performing microscopic level assembly.

\subsection{Interpreting the Prediction Accuracy Measure}\label{subsec:interpreting_pam}

The results presented in the next section compare different models using the prediction accuracy measure described in Section~\ref{subsec:prediction_accuracy_measure}. In this section, we present an interpretation of such a measure to provide an understanding of how the accuracy values map onto an error displacement measure. Equation~\ref{eq:pam} shows that the measurement is composed by two factors: i) linear and ii) angular displacement. The linear displacement is scaled with the respect of the size of the object, thus $|(p_e \textendash p_{gt})^\intercal D_S\|\in[0,\inf]$ represents a displacement proportional to the size of the object: $0$ means no displacement, $1$ means a displacement as large as the size of the object, $2$ a displacement twice as large as the object, and so on. In contrast, the angular distance $(1 \textendash {\langle q_e, q_{gt} \rangle}^2)\in[0,1]$ is bounded, since the maximum rotational distance in the Quaternion space is $180^{\circ}$ or $\pi$ radiants which correspond to a distance measure of $1$.

The prediction measure accuracy is hence not bounded, but it is a linear combination of two non-negative distances and it provides a possible range of values for the linear and angular distance. For example, a prediction accuracy of $0.2$ means that the linear distance cannot be larger than $\frac{1}{5}D_S$, thus for a $20\ cm$ cube it will be up to $4\ cm$ in all directions, and the angular distance cannot be larger than $\frac{\pi}{5}$ radiants or $36^{\circ}$.

\subsection{Training Set}\label{subsec:trainingset}

The training set is composed of a set of contact models for the manipulator and the environment, as described in sections~\ref{subsec:manipulator_contact_model} and~\ref{subsec:environment_contact_model}. Figure~\ref{fig:push_box_contacts} shows the two contact models used for the experiments. The first contact model refers to contacts when the objects has only flat surfaces and it was trained upon a cube with sides of length $20\ cm$. The second refers to objects that presents curved surfaces and it was trained upon a cylinder with height $20\ cm$ and radius $10\ cm$.

\subsection{Test Set}\label{subsec:testset}

Our test set is composed of six objects. The same cube and cylinder used in the training set are also part of the training set, however the physical parameters were sampled at each trial to create different contexts from the ones seen during the training. Three of the remaining four objects were a $10\ cm \times 20\ cm \times 20\ cm$ rectangular prism, a $30\ cm \times 20\ cm \times 20\ cm$ rectangular prism and a cylinder of height $20\ cm$ and radius $15\ cm$. The final object was a hybrid object with dimensions $20\ cm \times 20\ cm \times 20\ cm$ formed of half a cylinder and an isosceles triangular prism connected by their largest flat surfaces. Figure \ref{fig:objects} shows the test set. 

\section{Experimental Results}\label{sec:results}

This section presents our evaluation of the systems. First we evaluate the ability of our framework to select the correct contact and motion models at prediction time. We then evaluate the ability of our pose estimation module to estimate the pose of an object described as a PCOM. Finally we demonstrate the ability of our system's internal model to make prediction in novel contexts. Specifically we demonstrate that the unbiased predictors can make reliable predictions of how the object behaves under push operations, and that biased predictors can provide a better accuracy for specific environment/object pairs. 

\subsection{Contact Model Selection}\label{sec:results_selection}

We based our approach on the key idea that if we condition predictions upon local contacts we can achieve better generalisation. In this section we evaluate the ability of our system to identify the most similar initial contact models for conditioning the predictions of motion.

\begin{table*}[t]
\centering
\scriptsize
\caption{\footnotesize Experimental parameters for evaluating the selection of contact and motion models (Section \ref{sec:pose_estimation_accuracy}).}
\begin{tabular}{|l|l|l|l|}
    \hline
    \multicolumn{2}{|c|}{Contact Model Generation} & \multicolumn{2}{|c|}{Condition Generation} \\
    \hline
    Distance & $0.01$ & Number of Conditions to Generate & $50$ \\
    Lambda & $100$ & Number of Environment Contacts & $10$ \\
    & & Number of Samples When Generating Environment Contacts & $10$ \\
    \hline
    \multicolumn{2}{|c|}{Motion Model Training} & \multicolumn{2}{|c|}{Ground Truth Generation} \\
    \hline
    Number of Actions & $3$ & Number of Actions & $3$ \\
    Angle Range & $[-10,10]$ & Angle Range & $[-10,10]$ \\
    Action Duration ($s$) & $4$ & Action Duration ($s$) & $4$ \\
    Action Speed ($m s^{-1}$) & $0.1$ & Action Speed ($m s^{-1}$) & $0.1$ \\
    Samples Per Action & $500$ & Samples Per Action & $4$ \\
    Object Mass ($kg$) & $0.5$ & Object Mass ($kg$) & $0.5$ \\
    Object Coefficient of Friction & $U(0.15,0.35)$ & Object Coefficient of Friction & $U(0.15,0.35)$ \\
    Number of Environment Contacts & $10$ & & \\
    \hline
    \multicolumn{4}{|c|}{Prediction Generation} \\
    \hline
    \multicolumn{2}{|l|}{Number of Environment Contacts} & \multicolumn{2}{|l|}{$5$} \\
    \multicolumn{2}{|l|}{Environment Contact Kernels} & \multicolumn{2}{|l|}{$5000$} \\
    \multicolumn{2}{|l|}{Manipulator Contact Kernels} & \multicolumn{2}{|l|}{$500$} \\
    \hline
\end{tabular}
\label{tab:adaptive_model_experiment_parameters}
\end{table*}

We evaluate our trained models on a cube and a cylinder (Section~\ref{subsec:trainingset}) over three conditions. The \emph{congruent} condition refers to manually select the initial contact model so that the trained model is applied to the same object's shape it was train on (i.e. the contact model learned on a cube is applied on the same cube, and the contact model learned on a cylinder is applied to the same cylinder). The \emph{incongruent} condition refers to a manually mismatch so that the contact model trained on the cube is applied to a cylinder and viceversa. Finally, the \emph{adaptive} condition is when the system automatically selects the model using the local surface features as described in Section \ref{subsec:model_selection}.
Experimental parameters used for this experiment are given in Table \ref{tab:adaptive_model_experiment_parameters}.

\begin{figure}[H]
	\centering
	\includegraphics[width=0.9\columnwidth]{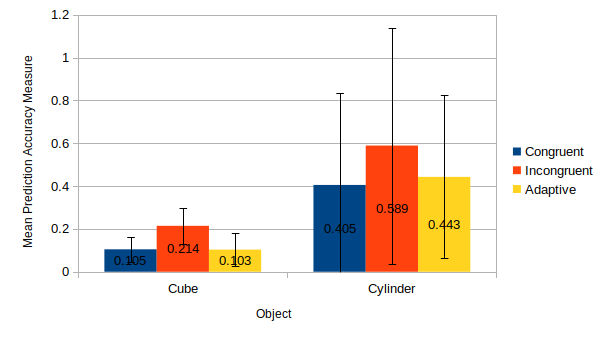}
	\caption{\footnotesize Mean prediction accuracy measures each over 50 pushes when applying congruent, incongruent and adaptive models to a cube and a cylinder.}
	\label{fig:adaptive_results}
\end{figure}

The results of the tests confirm our key idea. Figure \ref{fig:adaptive_results} shows that motion predictions on familiar ground (congruent and adaptive conditions) are better than when we force a mismatch (incongruent condition). It is important to note that in these experiments we do not test the ability of the framework to generalise to novel shapes, which will be presented in Sec~\ref{sec:results_accuracy}. However the main result is that the framework is capable to automatically select the most appropriate models, and prediction accuracy using the adaptive method is comparable to the case where the correct models are selecting by hand (i.e. congruent).

\subsection{Pose Estimation Accuracy} \label{sec:pose_estimation_accuracy}

\begin{table*}[t]
\centering
\scriptsize
\caption{\footnotesize Experimental parameters for the  experiments on pose estimation (Section \ref{sec:pose_estimation_accuracy})}
\begin{tabular}{|l|l|p{9cm}|}
\hline
\multicolumn{3}{|c|}{Pose Estimation} \\
\hline
Kernels & $3000$ & The number of kernels to be generated in the query density. \\[0.2cm]
Transform Standard Deviation Threshold & $5.0$ & Defines the transform distance cut-off for kernels in the query density in terms of a transform standard deviation multiplier. \\[0.1cm]
Principal Curvature Standard Deviation Threshold & $0.1$ & Defines the principal curvature distance cut-off for kernels in the query density in terms of a principal curvature standard deviation multiplier. \\[0.1cm]
Simulated Annealing Candidates & $500$ & The number of object position candidates to use when applying simulated annealing. \\[0.1cm]
Simulated Annealing Steps & $100$ & The number of steps to apply to each candidate during the simulated annealing process. \\[0.1cm]
Linear Kernel Bandwidth & $0.1$ & The drop-off rate of the query density kernels in terms of linear distance between two transforms. \\[0.1cm]
Angular Kernel Bandwidth & $20.0$ & The drop-off rate of the query density kernels in terms of angular distance between two transforms. \\
\hline
\end{tabular}
\label{tab:position_model_experiment_parameters}
\end{table*}

This experiments investigate the varying degrees of accuracy offered by using a position model as opposed to a baseline centroid based approach for approximating the position of an object. The experimental results have shown that this position model offers a significant increase in accuracy over the previously used centroid approach.

The experiments were carried out upon a cube with sides length $20\ cm$, a cylinder with height $20\ cm$ and radius $10\ cm$ and a modified cube. The modified cube is almost identical to the cube except with ${^{1}/_{16}}$ of the cube's volume removed from one of the corners. This 
\begin{figure}[H]
	\centering
	\includegraphics[width=0.99\columnwidth]{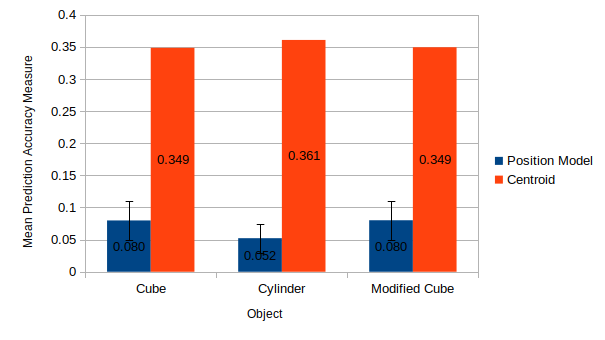}
	\caption{\footnotesize Results for object position estimation accuracy for position model and centroid based techniques. Mean prediction accuracy measures each over 100 runs is taken for the position model based technique meanwhile the output of the centroid technique is constant for a given input.}
	\label{fig:position_accuracy_results}
\end{figure}
\noindent removes some of the cube's symmetry and it is used to demonstrate the adaptability of the position model's approach.

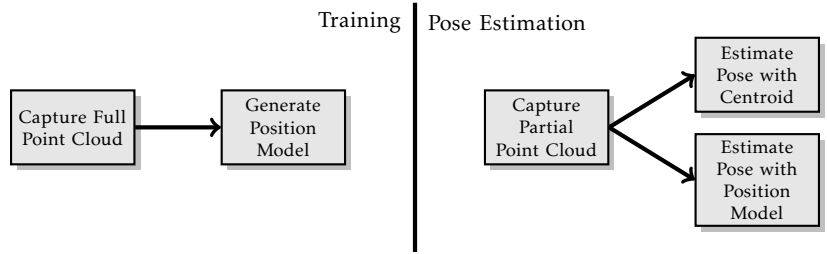
\begin{figure*}[t]
\centering
\begin{tikzpicture}[thick, scale=0.7, every node/.style={transform shape}]
\node (step1) [method-step] {Capture Full Point Cloud};
\path (step1)+(4,0) node (step2) [method-step] {Generate Position Model};
	
\path [draw, ->, ultra thick] (step1.east) -- node [above] {} (step2.west);

\path (step2)+(2.5,2.5) node (point1) {};
\path (step2)+(2.5,-2.5) node (point2) {};
\path (point1)+(-1.05,-0.5) node (training-text) {\large Training};
\path (point1)+(1.75,-0.5) node (training-text) {\large Pose Estimation};

\path [draw, ultra thick] (point1) -- node [above] {} (point2);

\path (step2)+(5,0) node (step3) [method-step] {Capture Partial Point Cloud};
\path (step3)+(4,1) node (step4) [method-step] {Estimate Pose with Centroid};
\path (step3)+(4,-1) node (step5) [method-step] {Estimate Pose with Position Model};

\path [draw, ->, ultra thick] (step3.east) -- node [above] {} (step4.west);
\path [draw, ->, ultra thick] (step3.east) -- node [above] {} (step5.west);
\end{tikzpicture}
\caption{\footnotesize Illustration of the methodology used for the pose estimation experiments.}
\label{fig:pose_accuracy_methodology}
\end{figure*}

Each object is experiment upon using the methodology illustrated in Figure \ref{fig:pose_accuracy_methodology}. The position model is run $100$ times and the mean prediction accuracy measure (see Sections \ref{subsec:prediction_accuracy_measure} and \ref{subsec:interpreting_pam}) is used to approximate the accuracy of our estimate for the initial pose of the object to be pushed. By contrast the centroid approach has its accuracy determined by the accuracy measure resulting from a single derivation of the point cloud centroid, as its output is always the same for a given input.\\
Experimental parameters used for this experiment are given in Table \ref{tab:position_model_experiment_parameters}.

The position model offers significant improvements over the centroid approach (see Figure \ref{fig:position_accuracy_results}). Application of the position model resulted a mean linear error of only $1.41\ cm$ (to 3 s.f.) as opposed to a linear error of $7.06\ cm$ for the centroid approach. Furthermore it is worth noting that the centroid approach cannot provide an estimate for the rotation of an object, whereas our position model can. As a testament to this, the worst accuracy measure derived for the position model throughout all $300$ applications was less than half the accuracy measure of the best accuracy measure associated with the centroid approach. Given that a lower accuracy measure corresponds directly to a higher level of accuracy, it can effectively be said that the position model's worst case performance has been demonstrated to be more than twice as accurate as the best performance of the baseline centroid approach.

\subsection{Prediction Accuracy on Novel Contexts}\label{sec:results_accuracy}

\begin{table*}[t]
\centering
\scriptsize
\caption{\footnotesize Experimental parameters for evaluating the prediction accuracy on novel objects (Section \ref{sec:results_accuracy}) under various friction biasing conditions.}
\begin{tabular}{|l|l|l|l|}
    \hline
    \multicolumn{2}{|c|}{Contact Model Generation} & \multicolumn{2}{|c|}{Condition Generation} \\
    \hline
    Distance & $0.01$ & Number of Conditions to Generate & $100$ \\
    Lambda & $100$ & Number of Environment Contacts & $5$ \\
    & & Number of Samples When Generating Environment Contacts & $100$ \\
    \hline
    \multicolumn{2}{|c|}{Motion Model Training} & \multicolumn{2}{|c|}{Ground Truth Generation} \\
    \hline
    Number of Actions & $3$ & Number of Actions & $3$ \\
    Angle Range & $[-10,10]$ & Angle Range & $[-10,10]$ \\
    Action Duration ($s$) & $4$ & Action Duration ($s$) & $4$ \\
    Action Speed ($m s^{-1}$) & $0.1$ & Action Speed ($m s^{-1}$) & $0.1$ \\
    Samples Per Action & $500$ & Samples Per Action & $3$ \\
    Object Mass ($kg$) & $N(0.5,0.025)$ & Object Mass ($kg$) & $N(0.5,0.025)$ \\
    Ground Plane Coefficient of Friction & $U(0.085,0.805)$ /  & Ground Plane Coefficient of Friction & $N(0.1,0.005)$ / \\
    & $N(0.1,0.005)$ /  & & $N(0.4,0.02)$ / \\
    & $N(0.4,0.02)$ /  & & $N(0.7,0.035)$ \\
    & $N(0.7,0.035)$ & & \\
    Number of Environment Contacts & $10$ & & \\
    \hline
    \multicolumn{4}{|c|}{Prediction Generation} \\
    \hline
    \multicolumn{2}{|l|}{Number of Environment Contacts} & \multicolumn{2}{|l|}{$5$} \\
    \multicolumn{2}{|l|}{Environment Contact Kernels} & \multicolumn{2}{|l|}{$5000$} \\
    \multicolumn{2}{|l|}{Manipulator Contact Kernels} & \multicolumn{2}{|l|}{$500$} \\
    \hline
\end{tabular}
\label{tab:friction_biasing_experiment_parameters}
\end{table*}

In this section we evaluate the generalisation abilities of our learned internal model for push operations. We demonstrate that both unbiased and biased predictors can be used to make predictions on how an object in previously unseen contexts behaves under a push operation. Unbiased predictors can be very useful when a good estimate for the physical parameters of the environment/object pair are unavailable, and in this section we will demonstrate that they are capable of providing a reliable ``guess'' for the test object's motion. Additionally, we will demonstrate that biased predictors offer a significant increase in the accuracy when some information about the environment/object context is available (e.g low friction/high mass).
The training and test test for these experiments are described respectively in Sections~\ref{subsec:trainingset} and~\ref{subsec:testset}.

In the experiments four conditions were considered to represent different variations of objects and environments as follows:
\paragraph{General} the model is trained to be unbiased. The parametrisation over the friction coefficient is represented as a uniform distribution over the unitless range $[0.085,0.805]$. For the mass distribution we use an uniform distribution over the range $[0.85,5.75]\ kg$.
\paragraph{Low} the model is trained to be either biased on a low friction coefficient or a low mass distribution. For friction, we employ a Gaussian distribution with mean $0.1$ and std dev $0.005$. For mass, we employ a Gaussian with mean $0.1\ kg$ and std dev $0.005\ kg$.
\paragraph{Medium} the model is trained to be either biased on a medium range friction or mass distribution. We employ respectively a Gaussian with mean $0.4$ and std dev $0.02$ and a Gaussian with mean $1.0\ kg$ and std dev $0.05\ kg$.
\paragraph{High} the model is trained to be either biased on a high range friction or mass distribution. We employ respectively a Gaussian with mean $0.7$ and std dev $0.035$ and a Gaussian with mean $5.0\ kg$ and std dev $0.25\ kg$.

Once predictions have been made for a set of push conditions for a given object and model the predictions are compared against the ground truths. The ground truths are the result of simulating the outcome under the push conditions for each of the setups relevant to the experiment taking place.\\
With that being said, the combination of two experiments for mass and friction, six objects, four model applications for each object and three setups for mass/friction results in a total of 144 different sets of simulated ground truths being compared with 48 sets of generated predictions.

Experimental parameters used for the experiment investigating biasing upon friction are given in Table \ref{tab:friction_biasing_experiment_parameters}. Meanwhile the experimental parameters used for the experiment investigating biasing upon mass are given in Table \ref{tab:mass_biasing_experiment_parameters}.

\begin{table*}[t]
\centering
\scriptsize
\caption{\footnotesize Experimental parameters for evaluating the prediction accuracy on novel objects (Section \ref{sec:results_accuracy}) under various mass biasing conditions.}
\begin{tabular}{|l|l|l|l|}
    \hline
    \multicolumn{2}{|c|}{Contact Model Generation} & \multicolumn{2}{|c|}{Condition Generation} \\
    \hline
    Distance & $0.01$ & Number of Conditions to Generate & $100$ \\
    Lambda & $100$ & Number of Environment Contacts & $5$ \\
    & & Number of Samples When Generating Environment Contacts & $100$ \\
    \hline
    \multicolumn{2}{|c|}{Motion Model Training} & \multicolumn{2}{|c|}{Ground Truth Generation} \\
    \hline
    Number of Actions & $3$ & Number of Actions & $3$ \\
    Angle Range & $[-10,10]$ & Angle Range & $[-10,10]$ \\
    Action Duration ($s$) & $4$ & Action Duration ($s$) & $4$ \\
    Action Speed ($m s^{-1}$) & $0.1$ & Action Speed ($m s^{-1}$) & $0.1$ \\
    Samples Per Action & $500$ & Samples Per Action & $3$ \\
    Object Mass ($kg$) & $U(0.085,5.75)$ / & Object Mass ($kg$) & $N(0.1,0.005)$ / \\
    & $N(0.1,0.005)$ / & & $N(1.0,0.05)$ / \\
    & $N(1.0,0.05)$ / & & $N(5.0,0.25)$ \\
    & $N(5.0,0.25)$ & & \\
    Ground Plane Coefficient of Friction & $N(0.3,0.05)$ & Ground Plane Coefficient of Friction & $N(0.3,0.05)$ \\
    Number of Environment Contacts & $10$ & & \\
    \hline
    \multicolumn{4}{|c|}{Prediction Generation} \\
    \hline
    \multicolumn{2}{|l|}{Number of Environment Contacts} & \multicolumn{2}{|l|}{$5$} \\
    \multicolumn{2}{|l|}{Environment Contact Kernels} & \multicolumn{2}{|l|}{$5000$} \\
    \multicolumn{2}{|l|}{Manipulator Contact Kernels} & \multicolumn{2}{|l|}{$500$} \\
    \hline
\end{tabular}
\label{tab:mass_biasing_experiment_parameters}
\end{table*}

\begin{figure*}[t]
\centering
\begin{subfigure}{0.9\columnwidth}
	\centering
	\includegraphics[width=\columnwidth]{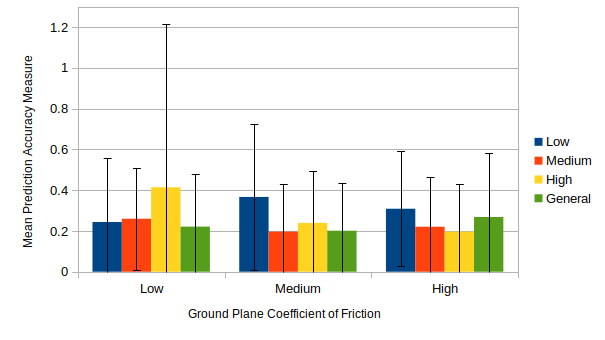}
	\justify Figure \arabic{figure}: \footnotesize Mean prediction accuracy measures each over 100 pushes produced by friction biasing experimentation for each model and condition (see Section~\ref{sec:results_accuracy}). The unbiased models (green) perform reasonably well in all conditions, although improvements can be achieved with biased predictors, e.g. high and medium models in high friction condition.
	\newcounter{friction_results}
	\addtocounter{friction_results}{\value{figure}}
\end{subfigure}
\stepcounter{figure}
\qquad
\begin{subfigure}{0.9\columnwidth}
	\centering
	\includegraphics[width=\columnwidth]{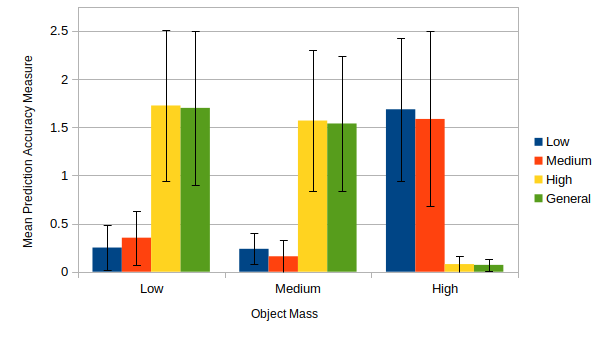}
	\justify Figure \arabic{figure}: \footnotesize Mean prediction accuracy measures each over 100 pushes produced by mass biasing experimentation for each model and condition (see Section~\ref{sec:results_accuracy}). The plot shows that the mass distribution has an higher impact on the predictions than the friction, and highlights more the need of biased models. Additionally it shows an unintentional bias of the general model (green) towards the high mass condition.
	\newcounter{mass_results}
	\addtocounter{mass_results}{\value{figure}}
\end{subfigure}
\end{figure*}

\begin{figure*}[t]
\centering
\begin{subfigure}{0.9\columnwidth}
	\centering
	\includegraphics[width=\columnwidth]{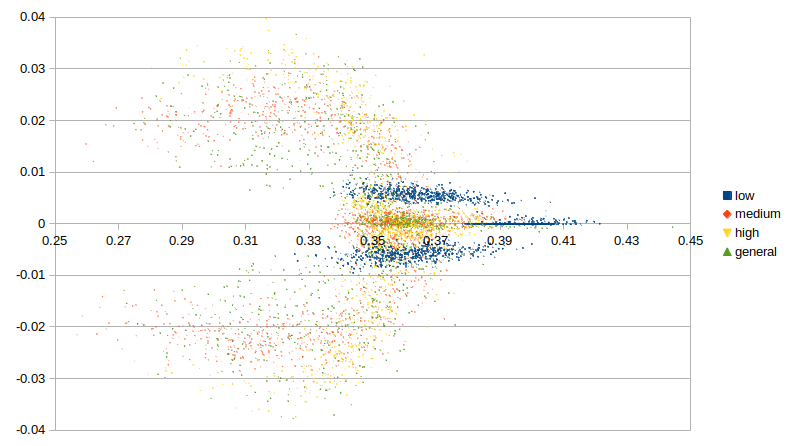}
	\justify Figure \arabic{figure}: \footnotesize 2D plot of the learned motion models for a cube in environments with different friction conditions (see Section~\ref{sec:results_accuracy}). The $x$ and $y$ axes in the plot are measured in metres. The $(0,0)$ pose represents the initial pose of the object to be pushed and each dot represents the pose of the object after a push in a given condition. The different distributions are best demarked by their colour coding. 
	\newcounter{friction_push_box_training_distributions}
	\addtocounter{friction_push_box_training_distributions}{\value{figure}}
\end{subfigure}
\stepcounter{figure}
\qquad
\begin{subfigure}{0.9\columnwidth}
	\centering
	\includegraphics[width=\columnwidth]{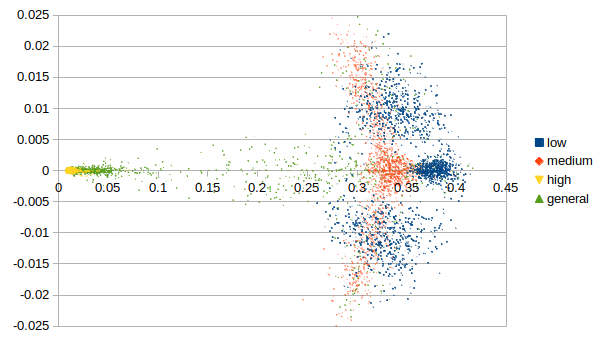}
	\justify Figure \arabic{figure}: \footnotesize 2D plot of the learned motion models for a cube in environments with different mass conditions (see Section~\ref{sec:results_accuracy}). The $x$ and $y$ axes in the plot are measured in metres. The $(0,0)$ pose represents the initial pose of the object to be pushed and each dot represents the pose of the object after a push in a given condition. The different distributions are best demarked by their colour coding.
	\newcounter{mass_push_box_training_distributions}
	\addtocounter{mass_push_box_training_distributions}{\value{figure}}
\end{subfigure}
\end{figure*}

\begin{figure*}[t]
\centering
\begin{subfigure}{0.9\columnwidth}
	\centering
	\includegraphics[width=\columnwidth]{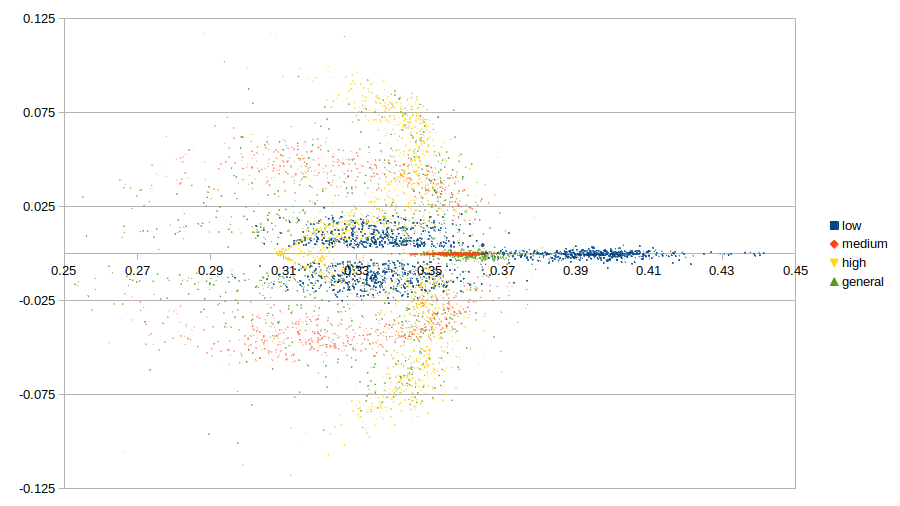}
	\justify Figure \arabic{figure}: \footnotesize 2D plot of the learned motion models for a cylinder in environments with different friction conditions (see Section~\ref{sec:results_accuracy}). The $x$ and $y$ axes in the plot are measured in metres. The $(0,0)$ pose represents the initial pose of the object to be pushed and each dot represents the pose of the object after a push in a given condition. The different distributions are best demarked by their colour coding.
	\newcounter{friction_cylinder_2_training_distributions}
	\addtocounter{friction_cylinder_2_training_distributions}{\value{figure}}
\end{subfigure}
\stepcounter{figure}
\qquad
\begin{subfigure}{0.9\columnwidth}
	\centering
	\includegraphics[width=\columnwidth]{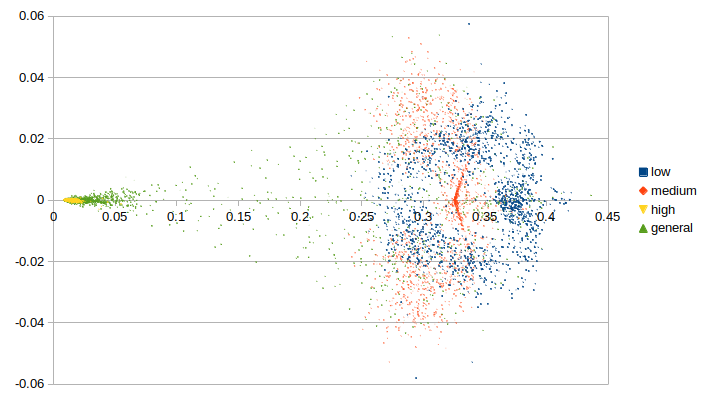}
	\justify Figure \arabic{figure}: \footnotesize 2D plot of the learned motion models for a cylinder in environments with different mass conditions (see Section~\ref{sec:results_accuracy}). The $x$ and $y$ axes in the plot are measured in metres. The $(0,0)$ pose represents the initial pose of the object to be pushed and each dot represents the pose of the object after a push in a given condition. The different distributions are best demarked by their colour coding.
	\newcounter{mass_cylinder_2_training_distributions}
	\addtocounter{mass_cylinder_2_training_distributions}{\value{figure}}
\end{subfigure}
\end{figure*}

The results of the friction biasing/generalising experiment demonstrate that models biased to a specific friction value predicted more accurately for cases with similar friction values and comparatively less accurately for other cases (See Figure \arabic{friction_results}). Furthermore, in situations where a set of models performed comparatively better they also possessed a smaller standard deviation in the prediction accuracy measure. This indicates that these biased models not only lead to better predictions on average for the situations upon which they were trained, but they also provide a greater level of reliability. 

Across most cases models performed better for ground plane frictions similar to the ones upon which they had been trained to (e.g. a high friction trained model performs better for medium friction cases than low friction cases). One exception to this is the case of applying a low friction trained model to high friction conditions yields better results than applying it to medium friction conditions. A likely reason for this can be seen by looking at the plots of final object positions during training in Figures \arabic{friction_push_box_training_distributions} \& \arabic{friction_cylinder_2_training_distributions}. One can see that the low friction distributions are tightly clustered and close to parts of the high friction distributions. Therefore for certain high ground plane friction situations the predictions given by the low friction model are going to match closely to the ground truth. Meanwhile the low friction training distributions are for the most part far away from the medium friction distributions. Hence, the low friction trained models perform comparatively better for high friction situations than medium friction situations. 

Of much greater interest however, is the behaviour of the generalised model, which closely mirror the friction biased model in the medium case. This is very unusual given that the unbiased model was trained on a broad range of friction values. We will discussed this issue in greater length at the end of this section.

The results of the evaluation upon mass closely mirror the results upon friction (See Figure \arabic{mass_results}). The low and medium mass conditions for the biased models perform similarly for all cases whilst the high mass biased model contrasts  the other models. Again, looking at the distributions of final object positions during training in Figures \arabic{mass_push_box_training_distributions} \& \arabic{mass_cylinder_2_training_distributions} it can be seen that the high mass distributions are very different to those generated by the low and medium mass cases. This again explains the disparity between the low and medium mass biased models and the high mass biased model. 

The generalised and high mass biased models perform similarly across the various cases. Further investigation revealed that this behaviour is a result of the underlying KDE method the models use. The various outcomes of push operations carried out during training form kernels which combine to create a PDF.

A generalised model attempts to provide predictions that generalise across a wide range of physical parameters. However, if certain physical parameters result in similar outcomes regardless of the push conditions (e.g. a high mass object which moves little) then this causes a large amount of kernels to be placed close to one another leading to a series of peaks representing a high likelihood for these positions during predictions. Hence when simulated annealing is carried out, predictions are all but guaranteed to come from these regions of high likelihood. Hence in cases where this occurs the generalised model has become unintentionally biased. As a result of this the model does not represent a true generalisation over that physical parameter. The risk of this occurring increases as the range of physical parameter values generalisation attempts to account for increases. Therefore as it stands the uniform range used for generalising models must be carefully selected in order to avoid this unintentional biasing.

\begin{figure}[t]
	\centering
	\includegraphics[width=\columnwidth]{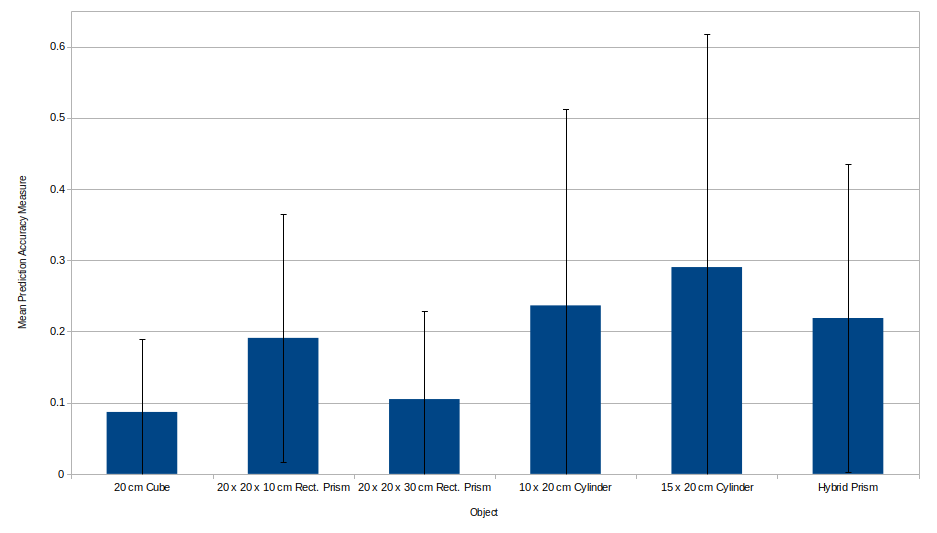}
	\caption{\footnotesize Mean prediction accuracy measures produced during biasing experimentation for each object. The mean is only taken for cases where the prediction time mass/friction conditions match biased training time conditions. This is to ensure that any difference in performance between objects is due to their shape incongruence and not factors relating to friction/mass incongruence or generalisation at training time. Since there is a low, medium and high biased model for both mass and friction, each with 100 test pushes, the mean prediction accuracy measure is taken over 600 test pushes. The three objects on the left of the figure were executed using a model trained on a $20\ cm$ cube. Meanwhile the three objects on the right of the figure were executed using model trained on a a cylinder with height $20\ cm$ and radius $10\ cm$. Objects tend to give performance similar to the object the model was trained with and the cylindrical objects generally perform worse, revealing a heightened difficulty of predicting for objects with more complex geometries. Finally, in the case of the $10\ cm$ cube, even though it appears to perform half as well as the $20\ cm$ cube upon which its model was trained, it is also half the size and therefore the linear part of its error is doubled, meaning the linear error may well likely be similar between the two.}
	\label{fig:object_results}
\end{figure}

Another aspect of the experiments that ought to be considered is how the various objects compared in their mean prediction accuracy measure. While it is true that models were only trained upon the $20\ cm$ cube and $20\ cm$ diameter cylinder it still provides some insight into the inherent difficulty for predicting for objects with different geometries. The cube provided the best prediction accuracy of all and has a raw linear error of only $1.68\ cm$, but this makes sense given it's symmetry in all dimensions and lack of curvature. The rectangular prisms performed comparatively worse, however, despite the disparity shown in Figure~\ref{fig:object_results} it is worth considering that the accuracy measure adjusts based upon the size of the object in question. Looking at the raw linear error associated with each rectangular prism, the $10\ cm$ prism has a linear error of $1.97\ cm$ while the $30\ cm$ prism has a linear error of $2.86\ cm$.\\

All of the objects possessing curvature performed comparatively worse. The $20\ cm$ and $30\ cm$ diameter cylinders had raw linear errors of $4.04\ cm$ and $8.14\ cm$ respectively. Both cylinders reflect a situation where a curved surface is in contact with the robot's manipulative link and a curved edge is in contact with the ground plane. The motions that result from the presence of these curves introduce additional complexity and this is exhibited in both the reduced prediction accuracy and the training distribution plots. The cylinder friction biasing distributions demonstrate a greater difference in distribution shape between friction conditions when compared to the friction biasing distributions of the cube. Meanwhile the cylinder mass biasing distributions exhibit a greater variation in the final object positions when compared with those of a cube. Both of these issues contributed towards a reduction in prediction accuracy, particularly the mass distributions which introduces a greater amount of variation in final object positions making the prediction problem inherently more difficult. Finally, in terms of the hybrid prism, the raw linear error was $3.96\ cm$. This is less error than the cylinder upon which the model was trained. However, this further evidences the increased difficulty of predicting for geometries involving curves, as while the robot's manipulative link is still in contact with a curved surface similar to the training cylinder, the rearmost edges at the base of the object are straight and hence the complexity of the interactions with the ground plane is reduced, as evidenced by the increased prediction accuracy.

\section{Conclusion \& Future Work}

This paper presents a model-based framework for learning transferable forward models for push manipulation. The model is constructed as a set of contact and motion models represented as probability density functions. The overall model is also parametrised over physical parameters which are critical for the task, e.g. mass and friction distributions. Our system behaves has an internal model which learns from experience physical interactions. In particular, we address in this work planar push interactions between a mobile robot, a 3D objects, and its environment. Our results show that our internal model can make reliable predictions in the presence of novelty in the object's shape and unknown physical parameters, efficiently transferring learned skills to novel contexts.    

In this work, it has become apparent that unbiased models tend to unintentionally bias during training. Although unbiased models still offer the capability of making reliable predictions without the need of fine tuning of the physical parameters, the main issue is a lack of information about the physical properties of the environment/object pair into the KDE kernel distance functions, which only rely on geometrical properties. Therefore a good direction for future work to pursue would be the integration of contact and motion models in a more compact representation. This compact representation should include an estimation of physical and geometrical properties in a single model, instead of the two separate. The main issue with this however, is that unlike principal curvatures or relational transform information the physical parameters cannot be derived at prediction time only from vision. 

Nonetheless, should it prove impossible to provide a good estimate by other means, it might be possible to derive a good estimate by carrying out several sample pushes in an online fashion. The friction and mass values could be refined and integrated into the models by comparing observed movements of the pushed object with the PDRs already available in the system. This would reflect how humans approximate these values, by observing the outcomes of physical interactions. 


%


\ifCLASSOPTIONcaptionsoff
  \newpage
\fi



%
%
%

\bibliographystyle{IEEEtran}
\bibliography{references}

%








\end{document}